\documentclass[twoside]{article}

\usepackage[round]{natbib} %

\usepackage{blindtext}
\usepackage{hyperref}
\usepackage{bbm}
\usepackage{mathtools}
\usepackage{booktabs}
\usepackage{amssymb,graphicx,url,amsmath, amsfonts}
\usepackage{times}
\usepackage{tikz}
\usepackage{soul}
\usepackage{color}
\usepackage{xcolor}
\usepackage{algorithm}
\usepackage{algorithmic}
\usepackage{enumerate}
\usepackage{comment}
\usepackage{amsthm}
\usepackage[makeroom]{cancel}
\usepackage{subcaption}
\usepackage{xparse}
\usepackage{balance}
\usepackage{url} 
\usepackage{dsfont}

\usepackage[accepted]{aistats2024}
\usetikzlibrary{shapes.geometric,positioning}

\newcommand{\Exp}{\mathbb{E}}

\newcommand{\cB}{{\cal B}}

\newcommand{\R}{{\mathbb R}}

\newcommand{\cM}{{\cal M}}

\newcommand{\bx}{{\bf x}}

\newcommand{\bn}{{\bf n}}

\newcommand{\bX}{{\bf X}}

\newcommand{\cF}{{\cal F}}

\newcommand{\ci}{{\mathfrak C}}

\newcommand{\bN}{{\bf N}}

\newcommand{\var}{{\rm Var}}

\newcommand{\PA}{{\bf PA}}
\newcommand{\pa}{{\bf pa}}
\newcommand{\abs}[1]{|#1|}
\newcommand{\cut}{\operatorname{cut}}

\newtheorem{Lemma}{Lemma}
\newtheorem{Definition}{Definition}
\newtheorem{Example}{Example}

\newcommand\independent{\protect\mathpalette{\protect\independenT}{\perp}}
\def\independenT#1#2{\mathrel{\rlap{$#1#2$}\mkern2mu{#1#2}}}

\usetikzlibrary{calc,arrows,shapes,plotmarks,positioning}
\usetikzlibrary{decorations.markings}

\tikzset{>=stealth'} 
\tikzstyle{graphnode} = 
 [circle,draw=black,minimum size=22pt,text centered,text
 width=22pt,inner sep=0pt] 
\tikzstyle{var} =[graphnode,fill=white]
\tikzstyle{vardashed} =[graphnode,draw=gray,fill=white]
\tikzstyle{obs} =[graphnode,fill=black,text=white]
\tikzstyle{obsgrey} =[graphnode,draw=white,fill=lightgray,text=black]
\tikzstyle{par} =[graphnode,draw=white,fill=red,text=black] 
 \tikzstyle{crucial} =[graphnode,draw=white,fill=yellow,text=black] 
\tikzstyle{fac} =[rectangle,draw=black,fill=black!25,minimum size=5pt]
\tikzstyle{facprior} =[rectangle,draw=black,fill=black,text=white,minimum size=5pt]
\tikzstyle{edge} =[draw=white,double=black,very thick,-]
\tikzstyle{blueedge} =[draw=white,double=blue,very thick,-]
\tikzstyle{rededge} =[draw=white,double=red,very thick,-]
\tikzstyle{prior} =[rectangle, draw=black, fill=black, minimum size=
5pt, inner sep=0pt]
\tikzstyle{dirprior} = [circle, draw=black, fill=black, minimum
size=5pt, inner sep=0pt]

\tikzstyle{dot_node}=[draw=black,fill=black,shape=circle]

\begin{document}

%
\runningtitle{Quantifying intrinsic causal contributions}

%
\runningauthor{Janzing, Bl\"obaum, Mastakouri, Faller, Minorics, Budhathoki}

\twocolumn[

\aistatstitle{Quantifying intrinsic causal contributions \\ via structure preserving interventions}

\aistatsauthor{ Dominik Janzing$^1$ \And Patrick Blöbaum$^1$   \And Atalanti A. Mastakouri$^1$   \AND Philipp M. Faller$^{1,2}$ \And Lenon Minorics$^1$ \And  Kailash Budhathoki$^1$}

\aistatsaddress{1 Amazon Research T\"ubingen, Germany
\And 2 Karlsruhe Institute of Technology, Germany }]

\begin{abstract}
We propose a notion of causal influence that describes the `intrinsic' part of the contribution of a node on a target node in a DAG. 
By recursively writing each node as a function of the upstream noise terms, we separate the intrinsic information added by each node from the one obtained from its ancestors. 
To interpret the intrinsic information as a {\it causal} contribution, we consider `structure-preserving interventions' that randomize each node in a way that mimics the usual dependence on the parents and does not perturb the observed joint distribution. To get a measure that is invariant with respect to relabelling nodes we use Shapley based symmetrization and show that it reduces  in the linear case to simple ANOVA after resolving the target node into noise variables. 
We describe our contribution analysis for variance and entropy, but contributions for other target metrics can be defined analogously.   
The code is available in the package gcm of the open source library \href{https://www.pywhy.org/dowhy/v0.11.1/user_guide/causal_tasks/quantify_causal_influence/icc.html}{DoWhy}. 
\end{abstract}

\section{INTRODUCTION}

Quantification of causal influence not only plays a role in expert's research on scientific problems, but also in highly controversial public discussions. For instance, the question to what extent environmental factors versus genetic disposition influence human intelligence, is an ongoing debate \citep{Krapohl2014}. 
Given the relevance of these questions, there is surprisingly little clarity about how to define strength of influence in the first place, see e.g. \cite{Rose2006}. 
More recent discussions on feature relevance quantification in explainable artificial intelligence have raised the problem of quantification of influence from a different perspective \citep{Datta2016,Lundberg2017,Frye2020,janzing2020feature,Shapley_Flow,Jung2022}. 
The quantification of causal influence proposed here is based on the intuition of measuring to what extent several factors influencing a target variable of interest `explain' the {\it variation} (aka uncertainty) of the latter. Attributing {\it uncertainty} of the target to the influencing factors  thus follows the spirit of  {\it Analysis of Variance} (ANOVA) \citep{Northcott}. 

Quantifying the causal influence has already resulted in a broad variety of proposals. 
A tentative taxonomy could classify measures according to the following three aspects:\\
\textbf{1/} What type of intervention is used:
 \citet{Ay_InfoFlow,Datta2016,Lundberg2017,Frye2020,janzing2020feature,heskes2020causal,Jung2022} use do-interventions  (also called point-interventions) on {\it nodes} of a causal directed acyclic graph (DAG) in the sense of \cite{Pearl:00}, while \cite{causalstrength,Schamberg2020} use interventions on the {\it edges}, and our paper uses {\it conditional} interventions. \\
\textbf{2/}  Which ones of the context variables are adjusted (and to which values) when the intervention is performed: Here and in most recent approaches one averages over all possible choices of  adjustment sets based on Shapley values (but most approaches adjust {\it values} of nodes, while we adjust their {\it mechanisms}). \\
\textbf{3/} Which target metric is used to assess the impact of those interventions: Here, we use {\it uncertainty} in the sense of entropy or variance, while 
  \citet{Datta2016,Lundberg2017,Frye2020,janzing2020feature,Shapley_Flow} use 
   shift of the {\it target value} itself or its {\it expectation} for an attribution referring to individual instances 
   at hand (as opposed to average contributions over populations). 
   As a further example,
    \citet{causalstrength,Schamberg2020} measure distribution change in terms of {\it relative entropy}.
    
Variations of these three different aspects can mostly be constructed independently in a modular manner. 
It is therefore pointless to blindly compare
measures of influence that differ by more than one aspect. To explain, for instance, why our paper uses {\it conditional} interventions instead of  {\it point} interventions,
we will compare it to hypothetical modifications of existing measures that would also be based on variance or entropy of a target node, rather than the target metric used in the respective literature.\footnote{We have seen that verbal discussions on causal attribution often blur the importance of {\it all three}  aspects, e.g. by claims like `we measure causal impact via Shapley values'.}

The crucial aspect of our proposal is what we consider the so-called `intrinsic' contribution. We explain this intuition via the following toy example which is paradigmatic for a large class of scheduling processes.
Consider the schedule of three trains $A, B, C$, where a delay in train $A$ causes a delay of  train $B$, which, in turn, causes a delay in train $C$. If we ask for the `intrinsic contribution' of $B$ to the delay of $C$, we are not asking for the hypothetical reduction of delay of $C$ if $B$ had arrived on time. Instead, we compare the delay of $C$ to the scenario where $B$ does not add any delay in addition to the one that it inherited from $A$. 
 In other words, the intrinsic contribution of $B$ on the delay of $C$ is obtained via a comparison with a scenario in which
 the dependences to the causal ancestors remain intact. Note that this distinction between `intrinsic' versus `inherited' part makes
sense regardless of whether we are  interested in the delay for a specific instance, the mean, or the variation of the delay of $C$ in the statistical population
(although this paper focuses on the latter). 


\paragraph{Probabilistic versus structural causal models}
While Causal Bayesian Networks \citep{Spirtes1993,Pearl:00} model causal relations via a DAG with random variables as nodes,   
a more `fine-grained' causal model is given by structural causal models:
\begin{Definition}[Structural Causal Model (SCM)]\label{def:fcm} 
An SCM corresponding to a DAG $G$ with observed variables $X_1,\dots,X_n$ as nodes is given by \\
(i) exogenous noise variables  $N_1,\dots,N_n$ being jointly statistically independent \\
(ii) functions $f_j$ that express each  $X_j$ deterministically in terms of its parents and the noise term, that is,
\begin{equation}\label{eq:fj}
X_j = f_j({\bf PA}_j,N_j).
\end{equation}
Moreover, \eqref{eq:fj} entails the counterfactual statement that, for any particular observation $(x_1,\dots,x_n)$, setting ${\bf PA}_j$ to ${\bf pa}_j'$ instead of ${\bf pa}_j$ 
would have changed $x_j$ to $x_j'=f_j({\bf pa}',n_j)$ (where $n_j$ denotes the value attained by $N_j$ for that particular statistical instance). 
\end{Definition}
Throughout the paper we leave it open whether or not the noise variables $N_j$  are observed or not. The idea is that $X_j$ are variables
whose values are more immediately accessible, while the $N_j$ or only its distribution $P(N_j)$ may be inferred from $X_1,\dots,X_n$.
Here and henceforth, we denote probability distributions by the capital letter $P$ and the corresponding densities or probability mass functions 
by $p$ (e.g. $P(X_1)$ versus $p(x_1)$). 

The existence of an SCM implies that $P(X_1,\dots,X_n)$ satisfies the Markov condition with respect to $G$ \citep{Pearl:00}.
On the other hand, every joint distribution that is Markovian relative to $G$ can be generated by an SCM, but this construction is not unique.
This is because knowing the causal DAG and the joint distribution alone does not determine all counterfactual causal statements (`rung 3 in the ladder of causation'  according to \citet{Pearl2018}).\footnote{See also \cite{causality_book}, Section 3.4, for an explicit description of the ambiguity.}

Section~\ref{sec:def} defines Intrinsic Causal Contribution (ICC) using the resolution into noise terms and quantifies the part of the uncertainty contributed by each noise variable. 
Section \ref{sec:inter} explains that ICC has a causal interpretation in terms of interventions on the variables $X_j$ although it requires adjustments of the noise at first glance. 
Then we discuss some properties of ICC in Section~\ref{sec:prop}. Section~\ref{sec:previous} compares ICC to related information based approaches in the literature. Finally, Section \ref{sec:exp} shows experiments with real data. 

We will see that the quantification of the influence of a variable on a target crucially depends on the type of interventions considered. Therefore, we argue that different notions of causal influence coexist for good reasons, since they formalize different ideas on what causal influence is about. 
We emphasize that this paper is mostly theoretical and its main contribution is conceptual, rather than containing new mathematical insights. 
Further, also questions on statistical robustness  and scalability for complex causal DAGs go beyond its scope. 
Constructing good proxies and adding assumptions that enable good estimates is left to future work, here we want to focus on providing 
an understanding of ICC to let readers judge themselves for which
 scenarios it is conceptually the right measure to ask for.

\section{DEFINING ICC }\label{sec:def}
\subsection{Resolving variables into noises} 

To discuss the idea of `intrinsic' within the language of SCMs,  consider first the following metaphoric example. 
An empty
donation box which starts at person $A$, is handed over to person $B$ and finally arrives at $C$. 
Each person adds the donation $N_A,N_B,N_C$, respectively.
If $X_j$ denotes the amount of money after the box passed person $j$, we 
obtain the simple SCM:
\begin{eqnarray}
X_A &=& N_A \\
X_B &=& X_A+ N_B \\
X_C &=& X_B + N_C,
\end{eqnarray} 
where we consider $N_j$  the  exogenous `noise' variables (which we know in this case), and the causal DAG reads
\begin{equation}\label{eq:donation} 
X_A \to X_B \to X_C.
\end{equation}
Obviously, we would consider $N_B$ the {\it contribution} of $B$ to the final amount $X_C$. Hence, the {\it noise terms}
carry the information about what has been added at the respective node $X_j$, not the values of the nodes $X_j$  directly. Clearly, $N_j$ captures what we would describe as one's {\it contribution} in our everyday language, not $X_j$. 
The example is metaphoric in the sense that it deals with contribution to a {\it fixed amount}, while we will later focus on contributions to the {\it uncertainty}.
We only want to motivate why our notion of contribution focuses on what is added by the {\it noise terms}. 
For this example one may argue that one could equally well define an augmented DAG 
containing $N_X$ explicitly as nodes in order to avoid focusing on {\it exogenous} variables. 
We argue, however, that the values $T_X$ are more directly observable, while the `contributions' $N_X$ are reconstructed 
from the former, using the above SCM. Likewise, the total delay of a train is {\it more directly observable} than
the amount contributed by the respective train itself, and we argue that similar statements apply to  a large class of problems of causal contribution analysis (see also Section \ref{sec:exp}), which
justifies our view of looking at contributions as something to be {\it reconstructed} from the SCM. 
 
To quantify the contribution of each ancestor to some target node $X_n$ of interest (which is assumed to have no descendants without loss of generality), we recursively insert structural equations \eqref{eq:fj} into each other and write $X_n$ entirely in terms of
the unobserved noise variables:
\vspace{-4pt}
\begin{equation}\label{eq:F}
X_n = F_n (N_1,\dots,N_n).
\end{equation}
Now we can think of $X_n$ as being the effect of the {\it independent} causes\footnote{Writing each node in terms of independent noise variables is also possible for 
some cyclic causal models, to which we could easily generalize ICC \citep{Bongers2021}.} $N_1,\dots,N_n$. 
Note that we do not introduce further assumptions regarding $F_n$, i.e. we allow arbitrary SCMs and do not limit it to a specific form such as additive noise models. 
While this resolution coincides with the root cause analysis of outliers by \citet{root_cause_analysis} and also 
with backtracking counterfactuals by \citet{backtracking}, ICC does not require the assumption of {\it invertible} SCMs, that is,
we do not need to be able to reconstruct the {\it values} of the noise from observed variables for single observations, we only need to know their distribution.  This is because 
ICC is not about contributions to single events.

\subsection{Quantifying conditional reduction of uncertainty} 

Next we quantify the reduction of uncertainty in $X_n$ caused by a hypothetical adjustment of $N_j$ (which coincides with usual conditioning 
due to exogeneity). 
\begin{Definition}[intrinsic causal contribution (ICC)] \label{def:cic}
Let $F_n$ as in \eqref{eq:F} express $X_n$ in terms of all noise variables $(N_1,\dots,N_n)=:\bN$. Then the ICC of
node $X_j$ (with $j=1,\dots,n$), given some additional adjustment set $T\subset \{1,\dots,n\}$, is defined by 
\begin{equation}
ICC_{\psi}(X_j\to X_n| T ) := \psi(X_n| \bN_T) - \psi(X_n| N_j, \bN_T) \label{eq:plainCIC},
\end{equation}
where $\psi$ can be any kind of conditional uncertainty measure satisfying {\bf monotonicity} $\psi(X_n | \bN_T) - \psi(X_n | N_j, \bN_T) \geq 0$ and
{\bf calibration} $\psi(X_n|\bN) =0$. Here, $\psi(.|\bN_T)$ denotes conditioning on all noise variables $N_j$ with $j\in T$. 
\end{Definition}
Monotonicity is not strictly needed. But in real-world applications positive contributions are easier to interpret and visualize (e.g., in pie charts), therefore we prefer a contribution that is non-negative.
Although the does not refer to interventions, Section \ref{sec:inter} explains that it is nevertheless causal because it can be recast in terms of (a non-standard type of) interventions, but we keep Definition \ref{def:cic} because of its simplicity.

Note that we decided to write $ICC_{\psi}(\cdot |T)$ instead of $ICC_{\psi}(\cdot| \bX_T)$ to emphasize that we do not condition on the random variables $\bX_T$, but the corresponding noise terms. 
Possible choices of $\psi$ are: 

\begin{Example}[conditional Shannon entropy]
With $\psi(X_n | \bN_T) := H(X_n | \bN_T)$ we obtain conditional mutual information \citep{cover}
\begin{align*}
ICC_{H}(X_j\to X_n| T ) & = H(X_n | \bN_T) - H(X_n | N_j, \bN_T)\\
& = I(N_j : X_n\,| \bN_T).
\end{align*}
Non-negativity of the r.h.s.
implies monotonicity of $\psi$. Calibration is satisfied for discrete $X_n$, while continuous entropy tends to minus infinity 
for a distribution approaching point measure. 
\end{Example}

We only use entropy when the target $X_n$ is discrete. 
Note that it does not matter whether the $N_j$ are continuous or discrete, in both cases we can measure the conditional uncertainty of  $X_n$  in terms of Shannon entropy.  


Although information theoretic quantification of influence comes with the advantage of being applicable to variables with arbitrary finite range, e.g., categorical variables, quantification in terms of variance is more intuitive and often easier to estimate from finite data: 
\begin{Example}[expected conditional variance] 
For $\psi(X_n | \bN_T) := \mathbb{E}[\var(X_n | \bN_T)]$ we obtain
\begin{eqnarray*}
&&ICC_{\var}(X_j\to X_n| T) =\\
&& \mathbb{E}[\var(X_n | \bN_T)] - \mathbb{E}[\var(X_n | N_j, \bN_T)].
\end{eqnarray*}
\end{Example}
The monotonicity of expected conditional variance follows from the law of total variance, $\var(Y) = \mathbb{E}[\var(Y \mid X)] + \var(\mathbb{E}[Y \mid X])$. Thus, $\mathbb{E}[\var(Y)] \geq \mathbb{E}[\var(Y \mid X)]$ (where $\mathbb{E}[\var(Y)]=\var(Y)$). Further, $\mathbb{E}[\var(Y)] \geq \mathbb{E}[\var(Y \mid X)]$, which entails also the conditional version  $\mathbb{E}[\var(Y\mid {\bf W})] \geq \mathbb{E}[\var(Y \mid {\bf W},X)]$ for any random vector
${\bf W}$.
Note that Analysis of Variance (ANOVA) relies on the same idea for the special case of a linear model where variances are just additive for independent factors \citep{Lewontin,Northcott}. 
Variance based sensitivity analysis \citep{Sobol2001} allows for non-linear models, but does not necessarily measure {\it causal} influence since 
it considers reduction of variance by conditioning on 
observed nodes regardless of whether the statistical relation to the target is causal or confounded.
In Section \ref{sec:inter} we explain a rephrasing of ICC in terms of interventions on observed nodes, which also admits
a generalization for semi-Markovian models.

\subsection{Symmetrization via Shapley values} 

Unfortunately, the contribution of each node $X_j$ in \eqref{eq:plainCIC} depends on the index subset $T\subset \{1,\dots,n\}$ given as context. 
For any ordering $\pi$ in the symmetric group $S_n$ we can certainly define `Plain ICC' by decomposing $\psi(X_n)$  into 
contributions $ICC_\psi (X_{\pi(j)}\to X_n|   T_\pi^j )$, where $T^j_\pi$ denotes the set of indices that occur before $j$ in the ordering $\pi$.  
Unfortunately, the dependence on $\pi$ introduces an undesired arbitrariness. 
Feature relevance quantification in explainable AI \citep{Datta2016,Lundberg2017,Frye2020,janzing2020feature} addresses similar problems via Shapley values from cooperative game theory \citep{Shapley1953}, which implicitly amounts to symmetrizing over all orderings $\pi$. 
For the definition of Shapley values see Section 1 in the appendix. 
\begin{Definition}[Shapley ICC] \label{def:Shapleycic}
Let the `worth of a coalition of noise terms $\bN_T$' for $T\subset \{1,\dots,n\}$ be given by $\nu(T):= -\psi(X_n|\bN_T)$. Then the (Shapley based) ICC of each node $X_j$ to
the uncertainty of $X_n$ reads:
\begin{eqnarray}
&&ICC_\psi^{Sh}(X_j \to X_n) \nonumber \\
&:=& \sum_{T\subseteq U\setminus \{j\}} \frac{1}{n {n-1 \choose |T|}} [ \nu(T\cup \{j\}) - \nu(T) ] \label{eq:shapley} \\
&=& \sum_{T\subseteq U\setminus \{j\}} \frac{1}{n {n-1 \choose |T|}} ICC_\psi(X_j\to X_n| T), \nonumber
\end{eqnarray} 
where $U:=\{1,\dots,n\}$
\end{Definition} 
Due to general properties of Shapley values (see Section \ref{sec:shapley}), Shapley based ICC values sum up to the uncertainty
of $X_n$:
 \begin{eqnarray}
&&\sum_{j=1}^n ICC_\psi^{Sh}(X_j \to X_n) = \nu(\{1,\dots,n\}) - \nu(\emptyset)  \nonumber\\
&=& \psi(X_n) - \psi(X_n| \bN) =\psi(X_n) . \label{eq:sumup} 
\end{eqnarray} 
Using an alternative equivalent definition of Shapley values, which enables simple Monte Carlo estimators \citep{Mitchel2022}, we can also average over $S_n$  instead:
\begin{equation}\label{eq:permbased}  
ICC_\psi^{Sh} = \frac{1}{n!} \sum_{\pi \in S_n} \psi(X_n| \bN_{T_\pi^j})-  \psi(X_n| \bN_{\{j\} \cup T_\pi^j}),
\end{equation} 
which we will also use later.

\begin{Example}[linear SCMs]\label{ex:linear} 
Given the linear SCM 
$\bX = A\bX + \bN$ where $\bX=(X_1,\dots,X_n)$ and $A$ is a lower triangular matrix.    Then, $\bX = (I - A)^{-1} \bN$ and thus $X_n$ is the linear combination 
of $n$ independent noise terms
$
X_n = \sum_{j=1}^n \beta_{nj} N_j,
$
where $\beta_{nj}$ is the last row of $(I-A)^{-1}$.  The contribution of each $N_j$ to the variance of $X_n$ thus
reads $\beta_{nj}^2 \cdot \var (N_j)$, independently
of the conditioning set $S$. Hence, we simply obtain   
\[
ICC^{Sh}_\var (X_j\to X_n) = \beta_{nj}^2 \cdot \var (N_j),
\]
which amounts to usual ANOVA (without the computational load entailed by Shapley values) after resolving $X_n$ into the noise terms.  
\end{Example}

\section{CAUSAL  INTERPRETATION}
\label{sec:inter}

\subsection{Structure-preserving interventions} 
We now argue that ICC can be rephrased using interventions on the nodes $X_1,\dots,X_n$: 
Going through the nodes in any topological ordering of the DAG, 
 {\it replace each $X_j$ with $f_j({\bf pa'}_j,n'_j)$ where $n_j'$ is a fixed value 
of an independent copy $N_j'$ of $N_j$}, and ${\bf pa'}_j$ denotes the values of the parents obtained 
from the upstream interventions. 
In contrast, standard interventions $do(X_j=x_j)$ replace structural assignment \eqref{eq:fj} with the trivial structural assignment $X_j:=x_j$. 
Although our intervention may be interpreted as setting the noise variable $N_j$ to $n_j'$ (which is unfeasible if we think of the exogenous noise of something that is not under our control, or even worse, not even observable), we suggest to interpret it as a intervention on $X_j$ instead that depends on the values of the parents. Given assignment \eqref{eq:fj}, we can intervene on $X_j$ without perturbing the joint distribution of all observable nodes $X_1,\dots,X_n$ as follows: after observing that ${\bf PA}_j$ attained the values ${\bf pa}_j$, we set $X_j$ to the value $x'_j:= f_j({\bf pa}_j,n_j')$ where $n_j'$ is randomly drawn from $P(N_j)$. Any statistical dependence between $X_i$ and our i.i.d. copy $N_j'$ of $N_j$ indicates causal impact of $X_j$ on $X_i$. This is because $N_j'$ is randomized and thus interventional and observational probabilities coincide, i.e. $P(\cdot | do(N_j' = n_j'))= P(\cdot | N_j'=n_j')$. 
Generalized interventions that replace the structural assignment \eqref{eq:fj} with a different one, have been studied in the literature earlier \citep{Eberhardt2007,Korb2004,Tian2001,Eaton2007,Markowetz2005,Correa_Bareinboim_2020}, but we keep  \eqref{eq:fj} and replace only $N_j$ with an {\it observable and accessible} copy $N_j'$. 

\subsection{Adjusting mechanisms at a node, not  their values}

The fact that we do not measure the reduction of uncertainty 
entailed by adjusting the {\it value} $x_j$ of the observed node $X_j$, but instead the 
reduction caused by adjusting the {\it noise}, can be nicely interpreted in terms of the response function formulation of SCMs \citep{Greenland1986}.
Then each value $n_j$ of $N_j$ corresponds to a function $f^{n_j}_j: {\bf pa_j}\mapsto x_j$ and the SCM is merely 
a probability distribution over the set of functions (`deterministic mechanisms') $f^{n_j}_j$. 
In this sense, ICC measures the extent to
which the variability of the respective mechanisms contribute to the variability of the target variable.   While phrasing ICC in terms of adjusting some of the $N_j$ resulted in a simple formal definition (to avoid distributions over functions), the more principled way of thinking of  ICC is  given by adjustments of the mechanism $f^{n_j}_j$. The drawing of values of the iid copy  $N_j'$ above now translates into iterating over randomized adjustments of the mechanisms.

Note that the decision of whether we are seeking for  a contribution measure that adjusts {\it values} or one that adjusts {\it mechanisms}  
is related to philosophical questions of blaming and 
the normative aspect of  `actual causation' \citep{Hitchcock2009,Halpern2013}. If we, for instance, blame
a train for its {\it entire} delay, we  implicitly consider the option of departing in time regardless of delays of `ancestor' trains (corresponding to adjusting the value), while 
intrinsic contribution accepts the dependence as mandatory by adjusting the mechanism.  

\subsection{Generalization of ICC to the confounded case \label{subsec:confound}}


Let us now assume a causal model with common causes of nodes $X_i,X_j$ which are not blocked by any set of nodes in $\{X_1,\dots,X_n\}$. 
The easiest way to generalize ICC to this case is given by allowing for
{\it dependent} noise terms.
We then define the worth of the coalition for any $T\subset \{1,\dots,n\}$ by  
\begin{eqnarray*} 
\nu(T)&:= &- \psi (X_n|do(\bN_T)) \\& :=& \left\{\begin{array}{c} \sum_{\bn_T}  H(X_n|do(\bN_T=\bn_T)) p(\bn_T) \\ 
\sum_{\bn_T}  \var (X_n|  do(\bN_T=\bn_T)) p(\bn_T)    \end{array}\right. ,
\end{eqnarray*} 
where the meaning of $do(N_T)$ is defined by the respective right hand sides. 
Note that in the unconfounded scenario with independent noise variables $N_j$ we did not explicitly introduce interventional probabilities since they 
coincided with conditional probabilities anyway. 
Assuming that the dependence of noise variables relies on common causes, the interventional probabilities are now given by the backdoor formula
\citep{Pearl:00} 
\begin{equation}\label{eq:interv}   
P(X_n|do(\bN_T=\bn_T)) = \sum_{\bn_T} P(X_n|\bn_{\bar{T}},\bn_T) p(\bn_{\bar{T}}),       
\end{equation}  
where $\bar{T}$ denotes the complement of $T$. 
Assuming that the noises are only connected by a common cause does not entail any loss of generality:
assume we had a path from one noise to another ($N_i \to N_j$), then we can introduce a mediator $N_i'$ between $N_i$ and $X_i$, and $N_i$ becomes a common cause of $N_i'$ and $N_j$. Since the only relevance of the noise consists in its effect on the response function ${\bf pa}_j \mapsto f_j({\bf pa}_j, n_i)$, the modified model responds in the same way to structure preserving interventions. 

For ease of notation, we have phrased $\nu(T)$ in terms of interventions on the  noise variables $N_j$ although we have emphasized in 
Section  3.2 in the main paper that ICC only requires interventions on the variables $X_j$, provided that we consider structure preserving interventions instead of usual do-interventions.
We can do the same here, but the interventions 
are now controlled by a simulated noise {\it vector} $\bN_T'$ which is a copy of $\bN_T$, that is, has the same {\it joint} distribution (rather than having separate copies $N'_j$  of each $N_j$). Note that we have $\nu(\{1,\dots,n\})= - \psi(X_n|do(\bN=\bn)) = 0$ and $\nu (\emptyset) =- \psi(X_n)$. Therefore,  Eq. (9) in the main paper holds for the
confounded case as well, which is crucial for the interpretation of attributing {\it the total uncertainty} of $X_n$ to its ancestors. 
Note, however, that ICC 
can become now negative because dropping a {\it correlated} noise variable can increase the variance of the target.


\section{SOME PROPERTIES OF ICC}\label{sec:prop} 

Here we describe properties that help getting an intuition on the behavior of ICC, particularly with respect to extending or marginalizing models. These properties will later help to understand the difference to other measures of causal influence in the literature and help the reader decide whether ICC is appropriate for the problem at hand. 

\subsection{Inserting nodes being perfect wires}\label{subsec:detinsert}
Assume we are given the causal DAG
$X \rightarrow Y$ 
with the structural equations
$X := N_X$ and $Y := f_Y(X,N_Y)$.
Then, application of Definition \ref{def:cic}, eq. \eqref{eq:plainCIC} and \eqref{eq:shapley} yields
\begin{eqnarray}\label{eq:ce}
ICC_\psi^{Sh} (X \to Y) &=& \frac{1}{2}[\psi(Y)- \psi(Y|N_X) \\
&+& \psi( Y|N_Y) - \psi( Y|N_X,N_Y)]. \nonumber
\end{eqnarray}
Let us now insert an intermediate node $\tilde{X}$ that is just an exact copy of $X$, that is, we define the modified SCM 
\begin{eqnarray}
X &=& N_X \label{eq:fcmchain1}\\ 
\tilde{X} &=& X \label{eq:fcmchain2}\\ 
Y &=& f_Y(\tilde{X},N_Y) \label{eq:fcmchain3}.
\end{eqnarray}
The corresponding DAG reads
$X \rightarrow \tilde{X} \rightarrow Y$.
From a physicists perspective, such a refinement of the description should always be possible because any causal influence propagates
via a signal that can be inspected right after it leaves the source.
Lemma \ref{lem:dummy} in Section \ref{sec:shapley} shows that \eqref{eq:fcmchain1} to \eqref{eq:fcmchain3} entail the same value for $ICC_\psi^{Sh}(X \to Y)$ 
as \eqref{eq:ce} because the constant `dummy' noise $N_{\tilde{X}}$ corresponding to $\tilde{X}$ is irrelevant for the contribution of the other nodes.


\subsection{Marginalizing over intermediate nodes or grandparents}\label{subsec:disSh}
While we inserted a deterministic node in Subsection~\ref{subsec:detinsert} we now marginalize over an intermediate node that depends non-deterministically on its cause.\footnote{Note that consistency of causal structures under various coarse-grainings is an interesting topic in a more general context too \citep{Rubensteinetal17}.}
Let us again consider the chain
\begin{equation}\label{eq:chain}
X \rightarrow Y \rightarrow Z,
\end{equation}
with the structural equations
\begin{eqnarray}
X &:=& N_X \label{eq:chain1}\\
Y&:= & f_Y(X,N_Y) \label{eq:chain2}\\
Z &:=& f_Z(Y,N_Z). \label{eq:chain3}.
\end{eqnarray}
Recall that, in case of entropy, $ICC_H^{Sh}(X\to Z)$ contains the terms $I(N_X: Z)$, $I(N_X:Z\,| N_Y)$, $I( N_X : Z\,| N_Z)$, $I(N_X: Z\,| N_Y,N_Z)$. 

Marginalizing over $Y$ yields the causal DAG $X\to Z$ with the structural equations
\begin{eqnarray}
X &:=& N_X \label{eq:chain1red}\\
Z &:=& \tilde{f}_Z(X,\tilde{N}_Z), \label{eq:chain2red}
\end{eqnarray}
where $\tilde{N}_Z := (N_Y,N_Z)$
and \[\tilde{f}_Z(X, \tilde{N}_Z) = f_Z(f_Y(X,N_Y),N_Z).\]
For the reduced structure, $ICC_H^{Sh}(X\to Z)$ contains only terms of the form $I(N_X: Z)$ and 
\[
I(N_X: Z\, |\tilde{N}_Z) = I(N_X: Z\,| N_Y,N_Z),
\]
while the terms $I(N_X:Z\,| N_Y)$, $I( N_X : Z\,| N_Z)$ do not occur any longer.
Hence, the Shapley based ICC is not invariant with respect to the marginalization. The reason is that Shapley symmetrization averages the relevance of $N_X$ over all possible combinations of background conditions. Reducing the possible combinations by ignoring nodes can result in different values for $ICC_\psi^{Sh}$. 
One may consider this as a caveat: we would not like the contribution of train A for the delay of train C depend on whether 
 B is explicitly taken into account or not. However, the reason is  a caveat that other Shapley based quantification of  feature relevance show as well:
Assume we are given $Y=f(X_1,X_2,X_3)$ and compute relevance of feature $X_1$ for $Y$ according to \cite{Lundberg2017}. One can then verify that the relevance 
of $X_1$ changes when we merge $X_2,X_3$ to a vector valued feature $X_2'$ and write $Y=f'(X_1,X_2')$. This is exactly what happens in our case:
marginalizing over $Y$ merges the noise variables $N_Y$ and $N_Z$  to one noise variable for $Z$.    

One also checks easily that the contribution of $Y$ changes when marginalizing over $X$. 
 In the limiting case where $Y$ is just a copy of $X$ we even obtain $ICC_\psi^{Sh}(Y\to Z)=0$ for DAG \eqref{eq:chain} while the DAG $Y\rightarrow Z$  is blind for the fact that
$Y$ has `inherited' all its information from its grandparent. Following our intuition of contribution, this change of ICC is required because 
we marginalization blurs that part of the uncertainty of $Y$ is just propagated from its parent.



\subsection{Continuity w.r.t. removing weak edges} 
To 
define a family of models for which an edge gets arbitrarily weak we consider variables attaining finitely many values and 
again use the response function formulation of SCMs, where 
 \eqref{eq:fj} turns into a probability distribution on the set $\cF_j$ of functions from the set of possible states of ${\bf PA}_j$ to the possible states of $X_j$. For each parent $PA^i_j$, there is a subset $\cF^i_j \subset F_j$ of functions that are constant in $PA^i_j$. 
`Fading away' of an edge from $PA^i_j$ can then be described by continuously turning the probability distribution into one with support $\cF^i_j$. 
For discrete variables, entropy and variance are continuous on finite sets, thus ICC will not show any discontinuity when the probability of the complement of
$\cF^i_j$ reaches  zero.

\subsection{Dependence on the SCM\label{subsec:rung3}}
ICC may differ for different SCMs (with the same DAG) describing the same interventional probabilities for interventions $do(X_i=x_i)$ (rung 3 versus rung 2 causal statements). As an example, let us take a look at $X\rightarrow Y$ with binary variables $X,Y$. First consider the structural equations
\begin{eqnarray}
X \coloneqq N_X, \quad Y \coloneqq X \oplus N_Y \label{eq:xor}
\end{eqnarray}
where $\oplus$ denotes the XOR operator. 
With $\psi=H$ and assuming `ùnbiased coins'' $P(N_X=1)=P(N_Y=1)=1/2$ we then obtain
\begin{eqnarray*}
ICC_H^{Sh}(X \to Y) &=& \frac{1}{2} (I(N_X:Y) + I(N_X: Y|N_Y) ) \\
&=& \frac{1}{2} ( 0 +1) =1/2.
\end{eqnarray*}
The same joint distribution $P(X,Y)$ can also be generated by 
\begin{eqnarray}
X \coloneqq N_X, \quad
Y \coloneqq N_Y \label{eq:cindep} 
\end{eqnarray}
for which we obtain\footnote{Readers may find it disturbing that \eqref{eq:xor} generates an {\it unfaithful} distribution and \eqref{eq:cindep} actually corresponds to
a DAG without arrows. But we have chosen this 
example only to keep the solution simple, dependence on SCM also holds for more generic models.} 
$ICC_H^{Sh}(X \to Y) = 0$. 
Hence, ICC also captures counterfactual influence which is invisible by usual interventions on the node itself. This can be desirable, for instance, in the following scenario:  
Let $X$ be a random variable denoting a text in binary encoding. Let $Y$ be its encrypted version generated by bitwise XOR with a randomly generated secret key $N$. 
If we have no access to $N$, we cannot detect any statistical dependence between $X$ and $Y$. However, we would not argue that the binary encoding of $X$ {\it did not contribute} to the encryption $Y$ just because the statistical dependence gets only visible after knowing $N$.
Even if we are not able to decrypt $Y$ because $N$ is unknown to us, the mere knowledge that $Y$ {\it could} be decrypted after knowing $N$ suffices to acknowledge the contribution of $X$ to $Y$.\footnote{The idea that noise terms $N_j$ in SCMs may be unobservable in one context, but not {\it in principle}, is also emphasized by \cite{Pearl:00} to justify the scientific content of counterfactuals.} 
As far as we can see, all known attempts based on usual point interventions fail to formalize our intuitive notion of `intrinsic', despite being reasonable concepts in their own right. This will be explained in the following section. 

\section{PREVIOUS WORK ON QUANTIFYING CAUSAL INFLUENCE}\label{sec:previous}

Since do-interventions are accepted as a crucial concept of causality, we want to check how they cold potentially capture `intrinsic' contribution without relying
on rung 3 causal models. 
To this end we consider the contribution of $X_2$ on $X_3$ in the causal chain 
\begin{equation}\label{eq:chain123} 
X_1\to X_2 \to X_3,
\end{equation} 
and emphasize that we want it to be zero when $X_2$ is a copy of $X_1$. We will see that most approaches fail in this regard, except for one case which
we reject because it violated continuity. 

\paragraph{Information Flow:} This measure, introduced by \citet{Ay_InfoFlow} and denoted by $I(X_2 \to X_3)$, measures the {\it mutual information} of $X_3$ and $X_2$ with respect to the joint distribution obtained when $X_2$ is subjected to randomized adjustments according to $P(X_2)$ (see Section \ref{sec:Ay} for the formal definition). This concept is certainly causal, but does not separate the information generated at $X_2$ from the one inherited from $X_1$. One option to achieve this separation would be to randomize $X_2$ according to the conditional distribution $P(X_2|do(X_1=x_1))$ instead of $P(X_2)$, which yields the {\it conditional} Information Flow $I(X_2 \to X_3| do(x_1))$ \citep{Ay_InfoFlow}. Its average over $P(X_1)$ is denoted by $I(X_2 \to X_3| do(X_1))$. For the example where $X_2$ is just a deterministic copy of $X_1$, we obtain $I(X_2 \to X_3| do(X_1))=0$, as desired, since adjusting $X_1$ sets $X_2$ to a constant. However, if $I(X_2 \to X_3| do(X_1))$ is a better candidate for 'intrinsic' contribution, generalization to arbitrary DAGs is unclear. 
The example suggests to measure the intrinsic contribution of any node $X_i$ in a DAG with $n$ nodes to the information on $X_j$ by  $I(X_i \to X_j\,| do(PA_i))$. 
Accordingly, in the DAG in Figure~\ref{left}, we would consider $I(X_2\to X_3\,| do(X_1))$ the intrinsic contribution of $X_2$ on $X_3$. 
\begin{figure}[h!]
\centering\scalebox{.8}{\begin{subfigure}[b]{0.5\columnwidth} 
 \begin{tikzpicture}[scale=0.55]
 \node[obs] at (0,2) (X_1) {$X_1$};
 \node[obs] at (2,2) (X_2) {$X_2$} edge[<-] (X_1);
 \node[obs] at (2,0) (X_3) {$X_3$} edge[<-] (X_1) edge[<-] (X_2); 
 \end{tikzpicture}
 \caption{}
 \label{left}
 \end{subfigure}
 \begin{subfigure}[b]{0.5\columnwidth} 
 \begin{tikzpicture}[scale=0.55]
 \node[obs] at (0,2) (X_1) {$X_1$};
 \node[obs] at (2,2) (X_2) {$X_2$};
 \node[obs] at (2,0) (X_3) {$X_3$} edge[<-] (X_1) edge[<-] (X_2); 
 \end{tikzpicture}
 \caption{}
 \label{right}
 \end{subfigure}
 }
\caption{\label{fig:triple} Left: Causal DAG for which it is already non-trivial to define the strength of the influence of $X_1$ on $X_3$ -- if one demands that this definition should also apply to the limiting case on the right (where the edge $X_1\to X_2$ disappeared).}
\end{figure} 
In Figure \ref{right}, $X_1$ is no longer a parent of $X_2$ and then we would consider choosing $I(X_2\to X_3)$ instead. 
This raises a conceptual problem raised by switching from conditional to unconditional Information Flow: let the edge $X_1\to X_2$ be arbitrarily weak, such that it disappears entirely, which entails a {\it discontinuous} change from $I(X_2\to X_3\,| do(X_1))$ to $I(X_2\to X_3)$. 
This is because the difference between $I(X_2\to X_3\,| do(X_1))$ and $I(X_2\to X_3)$ remains even when the edge disappears, which is easily seen for binary variables linked by a logical XOR gate: let $X_3 = X_1 \oplus X_2$ and $X_1$ be an unbiased coin. Without adjusting $X_1$, adjusting $X_2$ has no average influence on $X_3$, hence $I(X_2\to X_3)=0$. When adjusting $X_1$, however, $X_2$ controls $X_3$ entirely and when $X_2$ is unbiased too, we thus obtain $I(X_2\to X_3\,| do(X_1))=1 \,Bit$.

\paragraph{Causal Shapley Values and do-Shapley Values:}  
 Using the idea of these concepts \citep{heskes2020causal,Jung2022}, but also modifying it to {\it uncertainty} as our target metric,
 we define the worth of a coalition $T$ as $\psi(X_n| do(\bX_T)):= \sum_{\bx_T} \psi(X_n| do(\bX_T=\bx_T)) p(\bx_T)$.  
 Then 
 the contribution of $X_2$ on $X_3$ in the DAG \eqref{eq:chain123} is given by a sum of two terms of the form
 $\psi(X_n|do(\bX_T))   - \psi (X_n|do(X_2), do(\bX_T))$, which is non-zero for $T= \emptyset$ in contrast to what we demand.

\paragraph{Asymmetric Shapley Values:} 
Let us now modify \eqref{eq:permbased} by averaging only over all possible {\it topological orderings} of the DAG. 
For those $\pi$ we have 
\[
\psi(X_n| \bN_{T_\pi^j}) =  \psi(X_n|  \bX_{T_\pi^j}) ) = \psi(X_n|  do(\bX_{T_\pi^j})). 
\]
The first equality follows 
from $X_n \independent \bX_{T_\pi^j}\,| \bN_{T_\pi^j}$ and because
$ \bX_{T_\pi^j}$ is a function of   $\bN_{T_\pi^j}$.  The second one follows because conditioning on all ancestors blocks all backdoor paths. 
This way, we end up with a causal contribution measure that contains only observational conditionals and, more importantly, avoids rung 3 causal models. 
It turns out that this measure would be a straightforward modification of Asymmetric Shapley Values    
\citep{Frye2020}, with the only difference that we measure contribution to {\it uncertainty} in the population, while the former measures contribution 
of a specific value to the target value.  

Since the causal chain \eqref{eq:chain123} only admits one unique ordering, we obtain 
$\psi (X_3| do(X_2)) - \psi(X_3| do(X_2), do(X_1))$, which is zero when $X_2$ is a copy of $X_1$, as desired.    
Unfortunately, Asymmetric Shapley Values can change discontinuously when edges `fade away', because a DAG 
suddenly 
admits more topological orderings 
after removing an edge, which we wanted to avoid. For this reason, we believe that `fair' intrinsic causal contribution should account for all orderings, 
and thus also account for terms that result from adjusting mechanisms (aka noise variables) {\it downstream} while randomizing those of the {\it ancestors}
(which requires causal models of rung 3, see Subsection \ref{subsec:rung3}).  

Nevertheless, ICC with averaging over topological orderings only may be a reasonable approach, for instance in causal models of time series
without instantaneous effects. There, non-topological orderings would result in counterintuitive terms anyway which
randomize past noise variables, while adjusting later ones.

\paragraph{Strength of edges and paths:}  We sketch proposals for quantifying the strength of {\it edges} 
or {\it paths}   \citep{causalstrength,Shapley_Flow}    in Section  \ref{sec:edgespaths}.

\begin{figure*}[tb]
		\centering
		\begin{minipage}[c]{0.32\textwidth}
			\centering
			\includegraphics[width=\columnwidth, height=2.5cm]{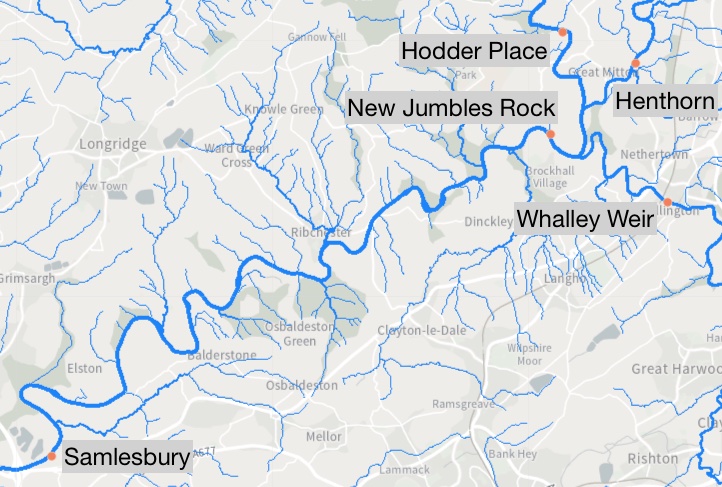}
		\end{minipage}
		\begin{minipage}[c]{0.33\textwidth}%
			\centering
			\includegraphics[width=\columnwidth, height=2cm]{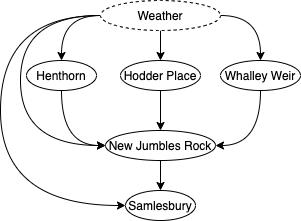}%
		\end{minipage}
		\begin{minipage}[c]{0.3\textwidth}%
			\centering
			\includegraphics[width=\columnwidth, height=3cm]{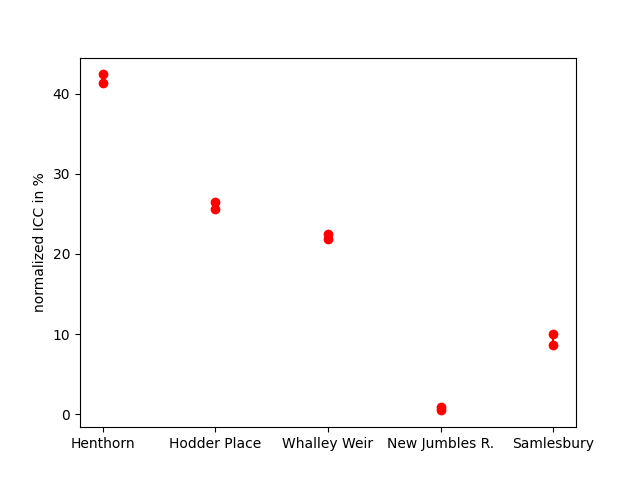}%
		\end{minipage}
	\caption{\label{fig:rivers} Left: location of the $5$ stations at which water flows are recorded. Middle: Causal model for the flows with Weather as latent confounder.
		Right: ICC with bootstrap confidence bounds for each of the $4$ upstream stations and the target station (Samlesbury) itself.} 
\end{figure*}

\section{EXPERIMENTS}\label{sec:exp}


\subsection{River flows}

We have used a data set with 3,953 daily records of the river flows (in $m^3/s$) at $5$ different stations in England at Henthorn,  Hodder Place, Whalley Weir, New Jumbles Rock, and Samlesbury\footnote{\url{https://environment.data.gov.uk/hydrology/explore}} between 1979/05/01 and 2021/12/11. Samleybury is considered the target since the other $4$ stations are upstream, as seen on the map on Figure \ref{fig:rivers}.
New Jumbles Rock lies at
a confluence point of the $3$ rivers passing Henthorn, Hodder Place, and Whalley Weir. The water passing a certain station is certainly a mixture of some fraction of the amount observed at the next stations further upstream plus some amount contributed by streams and little rivers  entering the river in between from the countryside around.
Again, the intrinsic contribution is the {\it unrecorded} amount required to explain the water flows observed at the respective station, given the flows 
at the adjacent stations upstream. Therefore, choosing the DAG (Figure~\ref{fig:rivers}, middle) containing only the observed flows seems more natural than an {\it augmented} DAG containing the
unrecorded influx as additional nodes, given that the latter values are reconstructed. 
Due to strong confounding effects (we have seen strong correlations of root nodes), estimating causal regression coefficients and noise by OLS would be heavily biased. Therefore we have inferred the SCM from common sense 
knowledge and set all regression coefficients to $1$,
assuming, for simplicity,  that most of the water flow at each node will also reach downstream nodes. Hence, the noise (i.e., the hidden influx) is simply the difference to the sum over all parents. Afterwards, we have applied ICC for the confounded scenario, see Subsection \ref{subsec:confound}, and Section \ref{sec:Appex} for more details.


Figure \ref{fig:rivers}, right, shows the ICC values. The low values for New Jumbles Rock and Samlesbury show that the flow at 
the root notes already explains most of the variation of the former. We have also applied  
variance based do-Shapley, as explained in Section \ref{sec:previous}, and obtained high contributions for 
New Jumbles Rock, see
Section \ref{sec:Appex}  for details.

\noindent
{\bf Adjusting  values versus adjusting mechanisms:}
The low value for New Jumbles Rock distinguishes ICC from causal influence measures that quantify contribution via the impact of a \emph{do}-intervention on the respective node: adjusting the flow at New Jumbles Rock reduces the variance of Samlesbury significantly. 
Assume politicians of Samlesbury want to reduce the fluctuations of water flows observed there (to enable safe shipping or reduce flooding risk) by 
building {\it retention dams} spread over the {\it countryside around} the rivers. Then the intrinsic contribution of each node is the maximal reduction achievable by dams in the respective region.  
A different strategy for getting more constant flow through Samlesbury would be to build dams {\it inside the river}, directly controlling the flow.  
Assume all the $5$ location were candidates for possible places. Trivially, having a dam in Samlesburry has the strongest impact, but there
may be other aspects for preferring one of the other $4$ locations. 
 In this case,
a measure of causal influence that quantifies the variance reduction by usual \emph{do}-interventions would be the right one to assess the impact of these dams. 
This example shows, again, that different measures of causal influence 
are appropriate for different goals.



\subsection{Fuel consumption of engines}

Here we consider 
a data set with non-linearities, as verified by scatter plots of 
bivariate dependences between some of the variables. Although we have still fitted an additive noise model as a convenient approximation, non-linearity of  structural equations entail that
then noise variables influence the target quantity in a non-additive way. 

The dataset AUTO MPG in the UCI Machine Learning Repository \citep{Bache2013}, contributed by R. Quinlan 
contain 7 features that are relevant
for the prediction of miles per gallon (mpg) of a car engine. The causal DAG given in  \citet{Wang2017}, after removing variables 
that have no influence on mpg, is shown in figure \ref{fig:mpg}. We have obtained
the following values for $ICC_{{\var}}^{Sh}$ after normalizing with the total variance:  {\tt cyl}  64\%, {\tt dis}  17\%, {\tt hp} $<1 \%$ , {\tt wgt} 3\%, {\tt mpg}  15\%. It turns out that the number of cylinders already explains a large fraction of the fuel consumption in the sense that the number commonly entails design decisions that involve parameters like the recorded ones. 
The intermediate nodes like {\tt dis}, {\tt hp}, and {\tt wgt} mostly inherit uncertainty from their parents, but roughly $1/7$ of the variance of {\tt mpg} remains unexplained by all of the above factors.

\begin{figure}
\centerline{
 		\scalebox{0.75}{\begin{tikzpicture}
 			\node[thick, draw, circle, font=\footnotesize, inner sep=1] (cyl) {cyl};
			 \node[thick, draw, circle, font=\footnotesize, right=1.5cm of cyl, inner sep=1] (dis) {dis};
			 \node[right=1.5cm of dis] (point) {};
 			\node[thick, draw, circle, font=\footnotesize, right=5cm of cyl, inner sep=1] (mpg) {mpg};
			\node[thick, draw, circle, font=\footnotesize, above=0.5cm of point, inner sep=1] (wgt) {wgt};
			\node[thick, draw, circle, font=\footnotesize, below=0.5cm of point , inner sep=1] (hp) {hp};
 			\draw[->, thick] (cyl) -- (dis);
			\draw[->, thick] (dis) -- (wgt);
			\draw[->, thick] (dis) -- (hp);
			\draw[->, thick] (hp) -- (mpg);
			\draw[->, thick] (wgt) -- (mpg);
 		\end{tikzpicture}}
		}
		\caption{\label{fig:mpg} Causal DAG of the AUTO MPG dataset for factors influencing fuel consumption with cyl: Cylinders, dis: displacement, hp: horsepower, wgt: weight, mpg: miles per gallon}
\end{figure}

\section{DISCUSSION}

\textbf{Computational complexity of ICC.}  The exact computation of the Shapley values is an expensive job which thus requires approximations. The package SHAP for explainable AI \citep{Lundberg2017} uses a probabilistic approach with random subsets (for a different kind of attribution though). We meanwhile work with a method that samples random permutations. This way, we simply replace the population average with an empirical mean, then the Hoeffding inequality yields error bounds that, a priori, do not depend on the number $n$ of nodes (but on bounds on information differences of subsets).
The choice of $f_j$ in~\eqref{eq:fj} and $\psi$ also influences the complexity of ICC. Example~\ref{ex:linear} shows a choice that simplifies computation. In particular, in linear models, variance based ICC does not involve costly averaging over subsets $S$. 
If we were to compute the reduction of variance entailed by {\it do-interventions} on the $X_j$ instead,
even the linear case would involve costly computation of Shapley values. 
 To see this, consider  the chain
 $
 X_1\to X_2 \to \cdots \to X_{n-1} \to X_n,
 $
 with the 
 structural equation $X_{j} = \alpha_{j-1} X_{j-1} + N_j$. The contribution of  $X_{n-1}$ on $X_n$ is computed via terms
 $\var (X_n|do(\bX_T))-  \var (X_n| do(X_{n-1}), do(\bX_T))$, and certainly the variance reduction by adjusting $X_{n-1}$ is smaller when 
 upstream nodes that are close by are already adjusted (such that the variance of $X_{n-1}$ is also low without adjusting it).  
 Thus, ICC raises less computational challenges than other Shapley value based measures for the linear case, and similar ones for non-linear models.


\textbf{Can we learn the SCM?} 
Although the observed conditionals $P_{X_j|{\bf PA}_j}$ do not in general determine the SCM, the hardness of inferring 
 $P_{X_j|{\bf PA}_j}$ from finite data often requires strong restriction of model classes such that the SCM follows `for free' \citep{Kano2003,Zhang_UAI,PetersMJS2014,Mooij2016}. For instance, additive noise model based inference \citep{Mooij2016} infers $P_{X_j|{\bf PA}_j}$ by fitting the structural equation $X_j = \tilde{f}_j({\bf PA}_j) + N_j$ with $N_j$ independent of ${\bf PA}_j$ and $\tilde{f}_j({\bf pa}_j):=\Exp[X_j|{\bf pa}_j]$. Hence, the SCM and also $N_j$ can then be entirely derived from observable entities due to $N_j=X_j - \Exp[X_j|{\bf PA}_j]$. For more SCM-learning methods see e.g. \cite{Storkey2006,Pawlowski2020}. 

\textbf{Does ICC reduce to some known measure after augmenting the DAG?} 
One may wonder whether ICC reduces to some known measure of causal influence in a DAG that contains all $X_j$ {\it and} $N_j$. However, in this augmented DAG usually interventions on $X_j$ and $N_j$ both can have an impact on the target. This way, we then obtain separate contributions for $X_j$ and for $N_j$\footnote{Here we assume that we apply Causal Shapley Values or do Shapley to all nodes in the augmented DAG, as usual. If we consider all variables $N_j$ as independent causes of $X_n$, then they coincide with ICC on this DAG.}, while ICC yields one contribution per $X_j$. 
In other words, no obvious modification of existing proposals seems to capture our notion of intrinsic by simply augmenting the DAG. 



\begin{thebibliography}{46}
\providecommand{\natexlab}[1]{#1}
\providecommand{\url}[1]{\texttt{#1}}
\expandafter\ifx\csname urlstyle\endcsname\relax
  \providecommand{\doi}[1]{doi: #1}\else
  \providecommand{\doi}{doi: \begingroup \urlstyle{rm}\Url}\fi

\bibitem[Ay and Polani(2008)]{Ay_InfoFlow}
N.~Ay and D.~Polani.
\newblock Information flows in causal networks.
\newblock \emph{Advances in Complex Systems}, 11\penalty0 (1):\penalty0 17--41,
  2008.

\bibitem[Bache and Lichman(2013)]{Bache2013}
K.~Bache and M.~Lichman.
\newblock {UCI} machine learning repository, 2013.
\newblock URL \url{http://archive.ics.uci.edu/ml}.

\bibitem[Blöbaum et~al.(2022)Blöbaum, Götz, Budhathoki, Mastakouri, and
  Janzing]{bloebaum2022dowhygcm}
P.~Blöbaum, P.~Götz, K.~Budhathoki, A.~A. Mastakouri, and D.~Janzing.
\newblock Dowhy-gcm: An extension of dowhy for causal inference in graphical
  causal models.
\newblock {\tt arXiv:2206.06821}, 2022.

\bibitem[Bongers et~al.(2021)Bongers, Forr{\'e}, Peters, and
  Mooij]{Bongers2021}
S.~Bongers, P.~Forr{\'e}, J.~Peters, and J.~M. Mooij.
\newblock Foundations of structural causal models with cycles and latent
  variables.
\newblock \emph{Annals of Statistics}, 49\penalty0 (5):\penalty0 2885--2915,
  2021.

\bibitem[Budhathoki et~al.(2022)Budhathoki, Minorics, Bloebaum, and
  Janzing]{root_cause_analysis}
K.~Budhathoki, L.~Minorics, P.~Bloebaum, and D.~Janzing.
\newblock Causal structure-based root cause analysis of outliers.
\newblock In K.~Chaudhuri, S.~Jegelka, L.~Song, C.~Szepesvari, G.~Niu, and
  S.~Sabato, editors, \emph{Proceedings of the 39th International Conference on
  Machine Learning}, volume 162 of \emph{Proceedings of Machine Learning
  Research}, pages 2357--2369. PMLR, 17--23 Jul 2022.

\bibitem[Correa and Bareinboim(2020)]{Correa_Bareinboim_2020}
J.~Correa and E.~Bareinboim.
\newblock A calculus for stochastic interventions:causal effect identification
  and surrogate experiments.
\newblock \emph{Proceedings of the AAAI Conference on Artificial Intelligence},
  34\penalty0 (06):\penalty0 10093--10100, Apr. 2020.

\bibitem[Cover and Thomas(1991)]{cover}
T.~Cover and J.~Thomas.
\newblock \emph{Elements of Information Theory}.
\newblock Wileys Series in Telecommunications, New York, 1991.

\bibitem[Daniel and Murphy(2007)]{Eaton2007}
E.~Daniel and K.~Murphy.
\newblock Exact bayesian structure learning from uncertain interventions.
\newblock In M.~M. and X.~Shen, editors, \emph{Proceedings of the Eleventh
  International Conference on Artificial Intelligence and Statistics}, volume~2
  of \emph{Proceedings of Machine Learning Research}, pages 107--114, San Juan,
  Puerto Rico, 2007. PMLR.

\bibitem[{Datta} et~al.(2016){Datta}, {Sen}, and {Zick}]{Datta2016}
A.~{Datta}, S.~{Sen}, and Y.~{Zick}.
\newblock Algorithmic transparency via quantitative input influence: Theory and
  experiments with learning systems.
\newblock In \emph{2016 IEEE Symposium on Security and Privacy (SP)}, pages
  598--617, 2016.

\bibitem[Eberhardt and Scheines(2007)]{Eberhardt2007}
F.~Eberhardt and R.~Scheines.
\newblock Interventions and causal inference.
\newblock \emph{Philosophy of Science}, 74\penalty0 (5):\penalty0 981--995,
  2007.

\bibitem[Frye et~al.(2020)Frye, Rowat, and Feige]{Frye2020}
C.~Frye, C.~Rowat, and I.~Feige.
\newblock Asymmetric shapley values: incorporating causal knowledge into
  model-agnostic explainability.
\newblock In H.~Larochelle, M.~Ranzato, R.~Hadsell, M.~Balcan, and H.~Lin,
  editors, \emph{Advances in Neural Information Processing Systems 33: Annual
  Conference on Neural Information Processing Systems 2020, NeurIPS 2020,
  December 6-12, 2020, virtual}, 2020.

\bibitem[Greenland and Robins(1986)]{Greenland1986}
S.~Greenland and J.~Robins.
\newblock Identifiability, exchangeability, and epidemiological confounding.
\newblock \emph{Int. Journal for Epidemiology}, 15\penalty0 (3):\penalty0
  413--9, 1986.

\bibitem[Halpern and Hitchcock(2013)]{Halpern2013}
J.~Halpern and C.~Hitchcock.
\newblock Graded causation and defaults.
\newblock \emph{The British Journal for the Philosophy of Science},
  66:\penalty0 413--457, 2013.

\bibitem[Heskes et~al.(2020)Heskes, Sijben, Bucur, and
  Claassen]{heskes2020causal}
T.~Heskes, E.~Sijben, I.~G. Bucur, and T.~Claassen.
\newblock Causal shapley values: Exploiting causal knowledge to explain
  individual predictions of complex models.
\newblock \emph{preprint {\tt arXiv:2011.01625}}, 2020.

\bibitem[Hitchcock and Knobe(2009)]{Hitchcock2009}
C.~Hitchcock and J.~Knobe.
\newblock Cause and norm.
\newblock \emph{The Journal of Philosophy}, 106:\penalty0 587--612, 2009.

\bibitem[Janzing et~al.(2013)Janzing, Balduzzi, Grosse-Wentrup, and
  Sch\"olkopf]{causalstrength}
D.~Janzing, D.~Balduzzi, M.~Grosse-Wentrup, and B.~Sch\"olkopf.
\newblock Quantifying causal influences.
\newblock \emph{Annals of Statistics}, 41\penalty0 (5):\penalty0 2324--2358,
  2013.

\bibitem[Janzing et~al.(2020)Janzing, Minorics, and
  Bloebaum]{janzing2020feature}
D.~Janzing, L.~Minorics, and P.~Bloebaum.
\newblock Feature relevance quantification in explainable ai: A causal problem.
\newblock In S.~Chiappa and R.~Calandra, editors, \emph{Proceedings of the
  Twenty Third International Conference on Artificial Intelligence and
  Statistics}, volume 108 of \emph{Proceedings of Machine Learning Research},
  pages 2907--2916, Online, 26--28 Aug 2020. PMLR.

\bibitem[Jung et~al.(2022)Jung, Kasiviswanathan, Janzing, Bl\"{o}baum, and
  Elias]{Jung2022}
Y.~Jung, S.~Kasiviswanathan, D.~Janzing, P.~Bl\"{o}baum, and B.~Elias.
\newblock $do$-shapley: Towards causal interpretation of model prediction.
\newblock In \emph{Proceedings of 39. International Conference on Machine
  Learning (ICML)}. 2022.

\bibitem[Kano and Shimizu(2003)]{Kano2003}
Y.~Kano and S.~Shimizu.
\newblock Causal inference using nonnormality.
\newblock In \emph{Proceedings of the International Symposium on Science of
  Modeling, the 30th Anniversary of the Information Criterion}, pages 261--270,
  Tokyo, Japan, 2003.

\bibitem[Korb et~al.(2004)Korb, Hope, Nicholson, and Axnick]{Korb2004}
K.~Korb, L.~Hope, A.~Nicholson, and K.~Axnick.
\newblock Varieties of causal intervention.
\newblock In C.~Zhang, G.~H., and W.~Yeap, editors, \emph{Trends in Artificial
  Intelligence}, volume 3157 of \emph{Lecture Notes in Computer Science}.
  Springer, 2004.

\bibitem[Krapohl et~al.(2014)Krapohl, Rimfeld, Shakeshaft, Trzaskowski,
  McMillan, Pingault, Asbury, Harlaar, Kovas, Dale, and Plomin]{Krapohl2014}
E.~Krapohl, K.~Rimfeld, N.~Shakeshaft, M.~Trzaskowski, A.~McMillan, J.-B.
  Pingault, K.~Asbury, N.~Harlaar, Y.~Kovas, P.~Dale, and R.~Plomin.
\newblock The high heritability of educational achievement reflects many
  genetically influenced traits, not just intelligence.
\newblock \emph{Proceedings of the National Academy of Sciences}, 111\penalty0
  (42):\penalty0 15273--15278, 2014.

\bibitem[Lewontin(1974)]{Lewontin}
R.~Lewontin.
\newblock Annotation: the analysis of variance and the analysis of causes.
\newblock \emph{American Journal Human Genetics}, 26\penalty0 (3):\penalty0
  400--411, 1974.

\bibitem[Lundberg and Lee(2017)]{Lundberg2017}
S.~Lundberg and S.~Lee.
\newblock A unified approach to interpreting model predictions.
\newblock In I.~Guyon, U.~V. Luxburg, S.~Bengio, H.~Wallach, R.~Fergus,
  S.~Vishwanathan, and R.~Garnett, editors, \emph{Advances in Neural
  Information Processing Systems 30}, pages 4765--4774. Curran Associates,
  Inc., 2017.

\bibitem[Markowetz et~al.(2005)Markowetz, Grossmann, and Spang]{Markowetz2005}
F.~Markowetz, S.~Grossmann, and R.~Spang.
\newblock Probabilistic soft interventions in conditional gaussian networks.
\newblock In \emph{Proceedings of the conference Artificial Intelligence and
  Statistics (AISTATS), PMLR R5:214-221}, 2005.

\bibitem[Mitchell et~al.(2022)Mitchell, Cooper, Frank, and Holmes]{Mitchel2022}
R.~Mitchell, J.~Cooper, E.~Frank, and G.~Holmes.
\newblock Sampling permutations for shapley value estimation.
\newblock \emph{Journal of Machine Learning Research}, 23\penalty0
  (43):\penalty0 1--46, 2022.

\bibitem[Mooij et~al.(2016)Mooij, Peters, Janzing, Zscheischler, and
  Sch\"olkopf]{Mooij2016}
J.~Mooij, J.~Peters, D.~Janzing, J.~Zscheischler, and B.~Sch\"olkopf.
\newblock Distinguishing cause from effect using observational data: methods
  and benchmarks.
\newblock \emph{Journal of Machine Learning Research}, 17\penalty0
  (32):\penalty0 1--102, 2016.

\bibitem[Northcott(2008)]{Northcott}
R.~Northcott.
\newblock {Can ANOVA measure causal strength?}
\newblock \emph{The Quaterly Review of Biology}, 83\penalty0 (1):\penalty0
  47--55, 2008.

\bibitem[Pawlowski et~al.(2020)Pawlowski, Coelho~de Castro, and
  Glocker]{Pawlowski2020}
N.~Pawlowski, D.~Coelho~de Castro, and B.~Glocker.
\newblock Deep structural causal models for tractable counterfactual inference.
\newblock In \emph{Proceedings of Neural Information Processing Systems
  (NeurIPS)}, 2020.

\bibitem[Pearl(2000)]{Pearl:00}
J.~Pearl.
\newblock \emph{Causality}.
\newblock Cambridge University Press, 2000.

\bibitem[Pearl(2001)]{Pearl_indirect}
J.~Pearl.
\newblock Direct and indirect effects.
\newblock In \emph{Proceedings of the Seventh Conference on Uncertainty in
  Artificial Intelligence (UAI)}, pages 411--420, San Francisco, CA, 2001.
  Morgan Kaufmann.

\bibitem[Pearl(2014)]{Pearl:2014}
J.~Pearl.
\newblock Interpretation and identification of causal mediation.
\newblock \emph{Psychological methods}, 19\penalty0 (4):\penalty0 459—481,
  2014.

\bibitem[Pearl and Mackenzie(2018)]{Pearl2018}
J.~Pearl and J.~Mackenzie.
\newblock \emph{The book of why}.
\newblock Basic Books, USA, 2018.

\bibitem[Peters et~al.(2014)Peters, Mooij, Janzing, and
  Sch{\"o}lkopf]{PetersMJS2014}
J.~Peters, J.~Mooij, D.~Janzing, and B.~Sch{\"o}lkopf.
\newblock Causal discovery with continuous additive noise models.
\newblock \emph{Journal of Machine Learning Research}, 15:\penalty0 2009--2053,
  2014.

\bibitem[Peters et~al.(2017)Peters, Janzing, and Sch\"olkopf]{causality_book}
J.~Peters, D.~Janzing, and B.~Sch\"olkopf.
\newblock \emph{Elements of Causal Inference -- Foundations and Learning
  Algorithms}.
\newblock MIT Press, 2017.

\bibitem[Rose(2006)]{Rose2006}
S.~P.~R. Rose.
\newblock Commentary: heritability estimates--long past their sell-by date.
\newblock \emph{International journal of epidemiology}, 35 3:\penalty0 525--7,
  2006.

\bibitem[Rubenstein et~al.(2017)Rubenstein, Weichwald, Bongers, Mooij, Janzing,
  Grosse-Wentrup, and Sch{\"o}lkopf]{Rubensteinetal17}
P.~K. Rubenstein, S.~Weichwald, S.~Bongers, J.~M. Mooij, D.~Janzing,
  M.~Grosse-Wentrup, and B.~Sch{\"o}lkopf.
\newblock Causal consistency of structural equation models.
\newblock In \emph{Proceedings of the Thirty-Third Conference on Uncertainty in
  Artificial Intelligence (UAI 2017)}, 2017.

\bibitem[Schamberg et~al.(2020)Schamberg, Chapman, Xie, and
  Coleman]{Schamberg2020}
G.~Schamberg, W.~Chapman, S.-P. Xie, and T.~P. Coleman.
\newblock Direct and indirect effects---an information theoretic perspective.
\newblock \emph{Entropy}, 22\penalty0 (8), 2020.

\bibitem[Shapley(1953)]{Shapley1953}
L.~Shapley.
\newblock A value for n-person games.
\newblock \emph{Contributions to the Theory of Games (AM-28)}, 2, 1953.

\bibitem[Sobol(2001)]{Sobol2001}
I.~Sobol.
\newblock Global sensitivity indices for nonlinear mathematical models and
  their monte carlo estimates.
\newblock \emph{Mathematics and Computers in Simulation}, 55\penalty0
  (1):\penalty0 271 -- 280, 2001.
\newblock The Second IMACS Seminar on Monte Carlo Methods.

\bibitem[Spirtes et~al.(1993)Spirtes, Glymour, and Scheines]{Spirtes1993}
P.~Spirtes, C.~Glymour, and R.~Scheines.
\newblock \emph{Causation, Prediction, and Search}.
\newblock Springer-Verlag, New York, NY, 1993.

\bibitem[Storkey et~al.(2006)Storkey, Simonotto, Whalley, Lawrie, Murray, and
  McGonigle]{Storkey2006}
A.~Storkey, E.~Simonotto, H.~Whalley, S.~Lawrie, L.~Murray, and G.~McGonigle.
\newblock Learning structural equation models for fmri.
\newblock In \emph{NeurIPS 2006}, 2006.

\bibitem[Tian and Pearl(2001)]{Tian2001}
J.~Tian and J.~Pearl.
\newblock Causal discovery from changes.
\newblock In \emph{Proceedings of the Seventeenth Conference on Uncertainty in
  Artificial Intelligence}, UAI’01, page 512–521. Morgan Kaufmann
  Publishers Inc., 2001.

\bibitem[Von~K\"ugelgen et~al.(2023)Von~K\"ugelgen, Mohamed, and
  Beckers]{backtracking}
J.~Von~K\"ugelgen, A.~Mohamed, and S.~Beckers.
\newblock Backtracking counterfactuals.
\newblock In M.~van~der Schaar, C.~Zhang, and D.~Janzing, editors,
  \emph{Proceedings of the Second Conference on Causal Learning and Reasoning},
  volume 213 of \emph{Proceedings of Machine Learning Research}, pages
  177--196. PMLR, 11--14 Apr 2023.

\bibitem[Wang and Mueller(2017)]{Wang2017}
J.~Wang and K.~Mueller.
\newblock Visual causality analysis made practical.
\newblock In \emph{IEEE Conference on Visual Analytics Science and Technology
  (VAST)}, pages 151--161, 2017.

\bibitem[Wang et~al.(2021)Wang, Wiens, and Lundberg]{Shapley_Flow}
J.~Wang, J.~Wiens, and S.~Lundberg.
\newblock Shapley flow: {A} graph-based approach to interpreting model
  predictions.
\newblock In A.~Banerjee and K.~Fukumizu, editors, \emph{The 24th International
  Conference on Artificial Intelligence and Statistics, {AISTATS} 2021, April
  13-15, 2021, Virtual Event}, volume 130 of \emph{Proceedings of Machine
  Learning Research}, pages 721--729. {PMLR}, 2021.

\bibitem[Zhang and Hyv\"arinen(2009)]{Zhang_UAI}
K.~Zhang and A.~Hyv\"arinen.
\newblock On the identifiability of the post-nonlinear causal model.
\newblock In \emph{Proceedings of the 25th Conference on Uncertainty in
  Artificial Intelligence}, Montreal, Canada, 2009.

\end{thebibliography}

\section*{Checklist}

 \begin{enumerate}

 \item For all models and algorithms presented, check if you include:
 \begin{enumerate}
   \item A clear description of the mathematical setting, assumptions, algorithm, and/or model: Yes
   \item An analysis of the properties and complexity (time, space, sample size) of any algorithm: No
 \end{enumerate}

 \item For any theoretical claim, check if you include:
 \begin{enumerate}
   \item Statements of the full set of assumptions of all theoretical results: Yes
   \item Complete proofs of all theoretical results: Yes
   \item Clear explanations of any assumptions: Yes     
 \end{enumerate}

 \item For all figures and tables that present empirical results, check if you include:
 \begin{enumerate}
   \item The code, data, and instructions needed to reproduce the main experimental results (either in the supplemental material or as a URL): Yes
   \item All the training details (e.g., data splits, hyperparameters, how they were chosen): Yes
         \item A clear definition of the specific measure or statistics and error bars (e.g., with respect to the random seed after running experiments multiple times): Not Applicable
         \item A description of the computing infrastructure used. (e.g., type of GPUs, internal cluster, or cloud provider): Not Applicable
 \end{enumerate}

 \item If you are using existing assets (e.g., code, data, models) or curating/releasing new assets, check if you include:
 \begin{enumerate}
   \item Citations of the creator If your work uses existing assets: Not Applicable
   \item The license information of the assets, if applicable: Yes
   \item New assets either in the supplemental material or as a URL, if applicable: Not Applicable
   \item Information about consent from data providers/curators: Not Applicable
   \item Discussion of sensible content if applicable, e.g., personally identifiable information or offensive content: Not Applicable
 \end{enumerate}

 \item If you used crowdsourcing or conducted research with human subjects, check if you include:
 \begin{enumerate}
   \item The full text of instructions given to participants and screenshots: Not Applicable
   \item Descriptions of potential participant risks, with links to Institutional Review Board (IRB) approvals if applicable: Not Applicable
   \item The estimated hourly wage paid to participants and the total amount spent on participant compensation: Not Applicable
 \end{enumerate}

 \end{enumerate}

\onecolumn
\appendix

\section{SHAPLEY VALUES \label{sec:shapley}}

\begin{Definition}[Shapley values]\label{def:sh}
Let $N$ be a set with $n$ elements (called `players' in the context of game theory) and $\nu : 2^N \rightarrow \R$ be a set function with
 $\nu(\emptyset) =0$ (assigning a `worth' to each `coalition').
Then the Shapley value of $i\in N$ is given by 
\begin{align}
\varphi_i(\nu) := 
\sum_{S \subset N \setminus \{i\}} \frac{ |S|! (n-|S|-1)! }{n!} ( \nu (S \cup \{i\}) - \nu (S) ) 
 = \sum_{S \subset N \setminus \{i\}} \frac{1}{n {n-1 \choose |S|}} ( \nu (S \cup \{i\}) - \nu (S) ). \label{eq:defSh}
\end{align}
\end{Definition} 
$\varphi_i(\nu)$ is thought of measuring the contribution of each player in a fair way and satisfies
\begin{equation}\label{eq:shadd}
\sum_{i=1}^n \varphi_i (\nu) = \nu(N).
\end{equation}

\begin{Lemma}[dummy noise variables]\label{lem:dummy}
Let $N_1,\dots,N_n$ be noise variables of an SCM $\cM$ with observed nodes $X_1,\dots,X_n$. Let $\tilde{\cM}$ be a modified SCM 
with observed variables $X_1,\dots,X_{n+k}$ and noise variables $N_1,\dots,N_{n+k}$ modeling the same 
joint distribution on $X_1,\dots,X_n,N_1,\dots,N_n$. Assume that the additional noise variables $N_{n+1},\dots,N_{n+k}$ are irrelevant for
$X_j$, that is 
\begin{equation}\label{eq:indN}
N_{n+1},\dots,N_{n+k} \independent X_j \, |\bN_T,
\end{equation}
for all $T \subset \{1,\dots,n\}$.
Then $\cM$ and $\tilde{\cM}$ yield the same values for $ICC_\psi^{Sh}(X_i \to X_j)$ for all $i=1,\dots,n$.
\end{Lemma}

We first need the following property of Shapley values: 

\begin{Lemma}[adding zero value players]\label{lem:zero}
For an extended set $\tilde{N}\supset N$, define a coalition function $\tilde{\nu} : 2^{\tilde{N}} \rightarrow \R$  by
\[
\tilde{\nu}(\tilde{S}) := \nu (\tilde{S} \cap N ),
\]
that is, $\tilde{\nu}$ is an extension of $\nu$ to irrelevant elements. Then
\[
\varphi_i (\tilde{\nu}) = \varphi_i (\nu) \quad \forall i \in N. 
\]
\end{Lemma}
\proof{It is sufficient to show the claim for the case where $\tilde{N}$ contains just one additional element, say $n+1$, since the remaining part follows by induction.

When computing $\varphi_i(\tilde{\nu})$ via a sum over all $\tilde{S}\subset \tilde{N}$ we can always merge two corresponding terms:
one set $S$ not containing $n+1$ and one corresponding set $S':= S \cup \{n+1\}$. 
Due to the irrelevance of $n+1$ we have

\begin{eqnarray*}
&&\tilde{\nu} (S' \cup \{i\}) - \tilde{\nu} (S') = \tilde{\nu} (S \cup \{i\}) - \tilde{\nu} (S) = \nu(S \cup \{i\} ) - \nu (S), 
\end{eqnarray*}

that is, both terms are the same as for the set $S$ in (9), up to the combinatorial factors. 
For the computation of $\tilde{\varphi}_i$, the term with $S$ comes with the factor
$ |S|! (n -|S|)! /(n+1)!$, while $S'$ comes with 
$(|S|+1)! (n -|S|-1)! /(n+1)!$.
The sum of these terms reads
\begin{eqnarray*}
&&\frac{|S|! (n-|S|-1)! }{(n+1)!} ( (n-|S|) + (|S|+1)) = \frac{ |S|! (n-|S|-1)!}{n!}, 
\end{eqnarray*}
which coincides with the factor in \eqref{eq:shapley}.$\square$} 

\newpage

To prove Lemma \ref{lem:dummy}, we note that 
$ICC_\psi^{Sh}$ is given by first defining the coalition function 
$
\nu (S) := -\psi(X_j | \bN_{S}),
$
for each $S\subset \{1,\dots,n\}$.
Then $ICC_\psi^{Sh}(X_i \to X_j) = \varphi_i(\nu)$. Further, define 
\begin{equation}\label{eq:indN2}
\tilde{\nu} (\tilde{S}) := -\psi(X_j | \bN_{\tilde{S}}).
\end{equation}
for each $\tilde{S}\subset \{1,\dots,n+k\}$.
Then,
$\tilde{\nu} (\tilde{S}) = \nu( \tilde{S} \cap \{1,\dots,n\})$. To see this, set $S:=\tilde{S} \cap \{1,\dots,n\}$. 
Then \eqref{eq:indN2} implies
\begin{equation*}
\psi(X_j | \bN_{\tilde{S}}) = \psi(X_j | \bN_{S}),
\end{equation*}
since conditioning on the dummy nodes does not change the distribution of $X_j$.
Hence, $\tilde{\nu}$ defines an extension of $\nu$ to irrelevant elements in the sense of Lemma~\ref{lem:zero}. 
Since $ICC_\psi^{Sh}(X_i\to X_j)$ with respect to the extended SCM is given by $\tilde{\varphi}_i (\tilde{\nu})$, the statement follows from Lemma~\ref{lem:zero}.

\section{STRENGTH OF EDGES AND PATHS \label{sec:edgespaths}}  

\subsection{Strength of causal arrows and indirect causal influence}  

\cite{causalstrength} defined 'strength of an edge' in the sense of an information theoretic quantity. 
It is based on an operation they called `cutting of edges'.
To quantify the information transferred along an arrow, one thinks of arrows as 'channels' that propagate information through space -- for instance `wires' that connect electrical devices. To measure the impact of an arrow
$X_j\to X_i$, they `cut' it and feed it with a random input that is an i.i.d.~copy of $X_j$. 
This results in the following `post-cutting' distribution:
\begin{Definition}[Single arrow post-cutting distribution] Let $G$ be a causal DAG with nodes $X_1,\dots,X_n$ and $P_{X_1,\dots,X_n}$ be Markovian with respect to $G$. Further, let $\PA^j_{i}$ denote the parents of $X_i$ without $X_j$.
Define the `post-cutting conditional' by 
\begin{align}\label{eq:postcut}
p_{X_j \rightarrow X_i}(x|\pa^{j}_i) := \sum_{x_j} p(x_i|\pa_{i}^{j}, x_j) p(x_j).
\end{align}
Then, the post-cutting distribution $P_{X_j \rightarrow X_i}(x_1,\dots ,x_n)$ is defined 
by replacing $p(x_i|\pa_i)$ in the causal factorization $p(x_1,\dots,x_n)= \prod_j p(x_j|\pa_j)$ with \eqref{eq:postcut}.
\end{Definition}

The relative entropy between the observed joint distribution $P(x_1, \dots, x_n)$ and the post-cutting distribution $P_{X_i \to X_j}$ now measures the strength of the arrow:
\[
\ci_{X_i\to X_j} := D(P\| P_{X_i\to X_j}).
\]
This measure is one of the concepts for which it is most apparent that its intention is different from ICC, and even complementary in a sense.
To see this, note that Postulate 2 in \cite{causalstrength} states that a measure of strength of an edge should be independent from how its tail node depends on its parents. This implies, in particular, that the values $\ci_{X_2\to X_3}$ for the DAGs 
$X_1 \to X_2 \to X_3$ and $X_2\to X_3$  coincide. Thus, the postulate explicitly requires $\ci_{X_i\to X_j}$ to ignore whether the information of $X_i$ has been inherited or not. 

This fundamental conceptual difference to ICC carries over to many other quantifications of causal influence, in particular the information theoretic {\it indirect} and {\it path-specific} causal influence in \cite{Schamberg2020}, which generalizes \citep{causalstrength}. For the quantification of the influence of $X_2$ on $X_3$ in $X_1\to X_2 \to X_3$, the indirect influence (which is the same as the direct one here) by \cite{Schamberg2020} coincides with the strength of the arrow $X_2\to X_3$ from \cite{causalstrength}. 
Further, note that also more classical approaches to mediation analysis and quantifying indirect effect, e.g., \cite{Pearl_indirect,Pearl:2014} have a different intention and do not distinguish whether the information node $X_2$ propagated to the target $X_3$ has been inherited from $X_2$'s parents or generated at the node itself.

\subsection{Shapley Flow}

\cite{Shapley_Flow} aim to quantify the contribution of {\it edges} to the value attained by a specific target node.
They argue that these attributions can also be used to measure the contribution of a {\it node} to the target.
First, we are going to explain the concept of Shapley Flow. 
In section \ref{subsec:icc_vs_sf} we want to argue, that ICC can be seen as a special case of a modified version of Shapley Flow.

\paragraph{Boundary consistency} 
\cite{Shapley_Flow} state that a quantification of the influence of edges must be invariant to \emph{refinement} of the causal model.
This is formalized in an axiom, which they call \emph{boundary consistency}.
They partition the nodes of a DAG $G$ into a data side $D$ and a model side $F$ (note that the terminology is motivated by applying their concept to model explainability in the context of explainable AI), where all root nodes (i.e. nodes without parents) of the DAG must lie in $D$ and the target node in $F$. 
Further, there must not be edges from $F$ to $D$.
The idea is that $F$ represents a new model with $D$ as input data.
Let $\phi_\cB(e)$ be the contribution of edge $e$ to the target node with respect to the boundary $\cB=(D, F)$. 
Let further $\cut(D, F)$ be the set of all edges with one end in $D$ and one end in $F$.
Then $\phi$ is boundary consistent, when 
\begin{equation}
	\phi_{\cB_1}(e) = \phi_{\cB_2}(e) \text{ iff } e\in (\cut(\cB_1) \cap\cut(\cB_2)).
\end{equation}

\paragraph{Update of edges} 
The goal of Shapley Flow is to attribute the change in the target node between a foreground setting $\bx$ and a background setting $\bx^*$ of the variables. 
To derive an attribution that fulfils the boundary consistency axiom, they traverse the DAG in a depth-first search (DFS) order and \emph{update} edges along the way.
We will discuss later, how the attribution is symmetrized over all possible DFS orders. 
When updated, an edge propagates an {\it updated} value (or the given foreground value for root nodes) to one child at a time.
The other children will be updated later, when they are explored by the DFS.
The child node uses this new value to update its own value according to the SCM (and stays at this value until updated again).
For its other parents the child resorts back to the last value it received in an update or their background values if the respective edge has not been updated yet.
This is motivated by the analogy to a message passing system that propagates the updated values step by step through the graph.
Formally, the \emph{history} of updates is represented as a list of edges.
The value of a variable $X_i\in D$ after updating the edges in history $h$ is calculated according to the SCM and the values of the parents $\PA_i$ that $X_i$ has been sent and stored, namely
\begin{equation}
	\label{eq:x_tilde}
	\tilde x_i(h) = f_i(m_{i,1}(h), \dots, m_{i,l}(h)), 
\end{equation} 
where $l=\abs{\PA_i}$ and $m_{i,j}(h)$ is the value in the memory slot where $X_i$ stores the value of its $j$-th parent node, after history $h$.
Let $\PA_{i,j}$ denote the $j$-th parent of $X_i$.
Initially (for the empty history), all these memory slots contain the respective baseline values from $\bx^*$, i.e. with $X_k = \PA_{i,j}$ we have
\begin{equation}
	m_{i,j}([\phantom{a}]) = x_k^*.
\end{equation}
Let $h$ be a history, $(X_k\to X_i)$ an edge and ${h' = h + (X_k\to X_i)}$ be the history that we get when we append $(X_k\to X_i)$ to the end of $h$.
When we already know the values $m_{i,j}(h)$ and $\tilde x_i(h)$ for $i=1, \dots, n$ and $j=1, \dots, \abs{\PA_i}$ we get the updated memory for $h'$ by changing the memory slot, where $X_i$ stores the value of $X_k$. 
This means, if $X_k=\PA_{i,j}$ we get
\begin{equation}
	m_{i,j}(h') = 
	\begin{cases}
		x_{k} &\text{ if } X_k \text{ is a root node}\\
		\tilde x_{k}(h) &\text{ else }
	\end{cases}
\end{equation}
where $x_k$ is the foreground value of $X_k$ from $\bx$ and $\tilde x_{k}(h)$ is the updated value of the node $X_k$ after history $h$ (as defined in equation \ref{eq:x_tilde}).
The other memory values stay the same, i.e.
\begin{equation}
	m_{\hat{i},j}(h') = m_{\hat{i},j}(h)
\end{equation}
for $\hat{i}=i$ and $\PA_{i,j}\neq X_k$ or $\hat{i}\neq i$ and $j=1, \dots,\abs{\PA_{\hat{i}}}$.
Note, that a node $X_k$ can have multiple children and can thus appear in the set $\PA_i$ for several $X_i$. 
Yet, the respective value does not have to be the same, since these children might have been sent an updated value of $X_k$ at different points along history $h$.
I.e., if $ \PA_{i,j} = X_k = \PA_{\hat i \hat j}$ for $i\neq\hat i$ that does not necessarily mean, that $m_{i,j}(h) = m_{\hat i \hat j}(h)$ for all $h$.

Once an update crosses the boundary $\cB=(D, F)$, it immediately reaches the target $Y$ over all paths, i.e. edges between two nodes in $F$ are assumed to instantly propagate the values of their parents. 
This reflects the idea that $F$ is thought of as a new model that operates normally, while Shapley Flow only alters the dependencies in the data.
Let for $X_i\in F$ the set $\PA_i^D$ be the parents of $X_i$ in $D$ and $\PA_i^F$ the parents in $F$.
Then replace the functions of the SCM according to 
\begin{equation}
	{f}^h_i(\pa_{i}^F) = f_i(\pa_i^F, \pa_i^D(h)).
\end{equation}
Note, that for a fix $h$ the value $\pa_i^D(h)$ is a constant.
As a special case, the equation of an $X_i$ does not change, if all its parents are in $F$.
The equations can recursively be solved for the variables in $D$ (like a normal SCM would be solved for the noise variables).
For a variable $X_i\in F$ the notation $\tilde x_i(h)$ refers to the value of $X_i$ after solving the functional equations for the nodes in $D$ and inserting the respective values for history $h$.

As an intermediate step, we define a function $\nu_\cB$ that assigns a value to a history, according to
\begin{equation}
	\nu_\cB(h) = \tilde y(h).
\end{equation}

\paragraph{Depth-first search}
The histories that Shapley Flow actually considers, are the ones that emerge from a DFS.
In a DFS, the graph is traversed in a recursive manner, i.e., each node recursively does a DFS on all its children.
Once the target $Y$ is reached, the algorithm returns.
Through this, the DFS finds all paths from the root nodes to the target node.
Note, that DFS does not dictate an order in which a node has to explore its children.
This means, there are several orderings in which a DFS can find the paths.
Further note, that the number of these possible orderings is smaller, than the number of all permutations of paths.
Once DFS is called on a node $X$, it will be called recursively on all children of $X$ before it backtracks .
Therefore, the respective paths always appear next to each other in an ordering derived from a DFS.

The purpose of these DFS orderings is to ensure boundary consistency.
To get some intuition, consider two boundaries $\cB_1=(D, F), \cB_2=(D\cup \{X\}, F\setminus\{X\})$, that only differ in one node $X\not\in D$.
Once a message passes the boundary, it is immediately propagated further by all downstream edges.
So, once an update reaches $X$ w.r.t. $\cB_1$, all outgoing edges of $X$ instantly propagate this update.
Once an update reaches $X$ w.r.t. $\cB_2$, all outgoing edges of $X$ will be updated, before any update can cross the boundary elsewhere.
This means for the other edges it \emph{looks like} nothing changed from $\cB_1$ to $\cB_2$.

\paragraph{Shapley contribution of edges}
The attribution to an edge $e$ is the sum of all path attributions to paths, that contain $e$.
This procedure can also be formulated as follows.
Let $\Pi_{DFS}$ be the set of orderings of paths, in which they can be found through depth-first search.
For path $i$ from root node to sink node define
\begin{align}\label{eq:sf}
	\tilde\phi(i) := \frac{1}{|\Pi_{DFS}|}\sum_{\pi\in \Pi_{DFS}} \big(\tilde\nu&\left(\left[ j\mid \pi\left( j\right) \le \pi \left( i\right)\right]\right) - \tilde\nu\left([j\mid \pi(j) < \pi(i)]\right)\big), 
\end{align}

where $\tilde \nu(S)$ operates on an list of lists of edges and evaluates the function $\nu_{\mathcal{B}}$ on the concatenation of all paths in $S$ and with respect to a boundary $\cB$.
Here we assume that a path is represented as a list of edges and that these lists of paths are sorted according to $\pi$.
Then we have
\begin{equation}
	\tilde{\nu}(S) := \nu_{\mathcal{B}}(s_1 + \dots + s_m), \quad m=\abs{S}, s_i\in S.
\end{equation}
Eventually, the Shapley Flow for an edge $e$ is
\begin{equation}
	\phi(e) := \sum_{p \in \text{ paths in } G} \mathds{1}[e \in p] \tilde\phi(p).
\end{equation}
 \cite{Shapley_Flow} implicitly assume that all relationships in the DAG are deterministic, with only the root nodes being random.
Put into the framework defined in Eq. (1) in the main paper, this means, that all nodes depend either on endogenous parents or on exogenous noise, i.e., all structural equations have one of the following forms
\begin{equation}
	X_j = f_j(\PA_j) \quad \text{ or } \quad X_j = f_j(N_j).
\end{equation}
If one wishes to calculate the Shapley Flow on a general SCM, the \emph{augmented} DAG is needed, containing all the noise variables $N_j$ as additional nodes with the edges ${N_j \to X_j}$.

\subsection{Describing ICC using modified Shapley Flow}
\label{subsec:icc_vs_sf}

\paragraph{Adding the noise variables} 
Now consider the augmented DAG $G'$ with $2n$ nodes and the boundary $\cB' := (\bN, \bX)$.
Since the SCM assumes causal sufficiency (i.e. all noise terms are only causing one variable) the cut of our boundary $\cB'$ contains the $n$ edges $N_j\to X_j$.
Further, the set of DFS paths from the nodes $N_j$ to the boundary is also $n$ (namely the paths $N_j \to X_j$) and the number of DFS orderings thereof is $n!$.
\cite{Shapley_Flow} emphasize that in general attributions to nodes can be made by summing up the attribution to all outgoing edges.
Since the nodes $N_j$ only have $N_j \to X_j$ as outgoing edges, we can interpret their contribution as the contribution of $X_j$. 
We will therefore identify the edges with their respective noise node in the following equation.
Rephrasing Equation \eqref{eq:sf} with respect to boundary $\cB'$ results in 
\begin{align*}
	\tilde\phi(i) = &\frac{1}{n!}\sum_{\pi\in\Pi} [\tilde\nu\left(\left\{ j\mid \pi\left( j\right) \le \pi \left( i\right)\right\}\right) - \tilde\nu\left(\{j\mid \pi(j) < \pi(i)\}\right)]
	 = \sum_{S \subset N \setminus \{i\}} \frac{1}{n {n-1 \choose |S|}} ( \tilde\nu (S \cup \{i\}) - \tilde\nu (S) ),
\end{align*}
 where $\Pi$ is the set of all permutations over $N_1,\dots,N_n$.
Hence, the Shapley Flow amounts to measuring the impact of updating the edges $N_j \to X_j$ in a symmetrized way. 
This way, it reduces to the standard way of measuring the impact of $n$ independent causes $N_1,\dots,N_n$ on the value of the target node. 
 
\paragraph{Replacing the target metric} Since Shapley Flow quantifies contribution of edges to the {\it value} of a target node, while we want to measure the uncertainty of the target node, we need a different choice for the set function $\tilde{\nu}$: instead of measuring the value of the target node, we need a metric that measures the uncertainty of the target
after activating an edge by feeding with an independent random input. This amounts to replacing the set function above with an uncertainty measure $\psi$ to finally obtain $ICC$. 
 
\paragraph{Remark} With this reinterpretation, the flow conservation property of Shapley Flow nicely captures the intuition that the strengths of the outgoing edges sum up to the flow coming from the parents plus the intrinsic contribution. This is seen as follows.
The flow property implies
 \begin{equation}
	\sum_{X_j\in \PA'_i} \phi(X_j\to X_i) = \sum_{X_k\in {\bf CH}'_i}\phi(X_i\to X_k),
\end{equation}
with $\PA'_i$ and ${\bf CH}'_i$ being the parents and children, respectively, of $X_i$ in the augmented DAG $G'$. This amounts to
\begin{eqnarray*}
	&& \sum_{X_j\in \PA_i} \phi(X_j\to X_i) + \phi(N_i\to X_i) 
	= \sum_{X_k\in {\bf CH}_i}\phi(X_i\to X_k),
\end{eqnarray*}
where $\PA_i$ and ${\bf CH}_i$ are the parents and children, respectively, of $X_i$ in the original DAG $G$, and $\phi(N_i\to X_i)$ measuring the intrinsic contribution.

\section{DEFINITION OF INFORMATION FLOW \label{sec:Ay}} 

As pointed out by \cite{Ay_InfoFlow}, quantifying causal influence between observed nodes via the information they share, requires computing information with respect to {\it interventional} probabilities rather than information given in a passive {\it observational} scenario (recall that this distinction has been irrelevant for us since dependences between observed nodes and noise are always causal and unconfounded). Accordingly, they define the {\bf information flow} from a set $\bX_A$ of variables to another set $\bX_B$, imposing that some background variables $\bX_S$ are set to $\bx_S$ by 
\begin{align} \label{eq:if}
&I(\bX_A \to \bX_B \,| do(\bx_S)):= - \sum_{\bx_A,\bx_B} p(\bx_B| do(\bx_A,\bx_S)) p(\bx_A|\bx_S) \log \frac{p(\bx_B|do(\bx_A,\bx_S))}{\sum_{\bx_A'} p(\bx_B|do(\bx_A', \bx_S)) p(\bx_A'|do(\bx_S))}.
\end{align}
Here, $p(\cdot | do(\bx_A, \bx_S))$ is the interventional probability after setting 
$\bX_A, \bX_S$ to $\bx_A, \bx_S$.
Moreover, \cite{Ay_InfoFlow} define also the average of \eqref{eq:if} over all $\bx_S$: 
\begin{eqnarray}
&&I(\bX_A \to \bX_B \,| do(\bX_S)):= \sum_{\bx_S} I(\bX_A \to \bX_B \,| do(\bx_S)) p(\bx_S).\nonumber
\end{eqnarray}

Note that $I(\bX_A \to \bX_B \,| do(\bx_S))$ measures the mutual information of $\bX_A$ and $\bX_B$ when $\bX_S$ is set to $\bx_S$ and
$\bX_A$ is randomized with probability $p(\bX_A|do(\bx_S))$.

\section{ADDITIONAL INFORMATION EXPERIMENTS AND MORE DETAILS \label{sec:Appex}}

For ICC on the Auto MPG data set we have used the publicly available implementation of ICC in \href{https://www.pywhy.org/dowhy/v0.11.1/user_guide/causal_tasks/quantify_causal_influence/icc.html}{DoWhy} \citep{bloebaum2022dowhygcm}, where we have chosen the auto-assign function. 

For the river flows, we have worked with an SCM where each variable is a sum of its parents plus noise, that is,
all regression coefficients are set to $1$. Accordingly, the variance of the target after adjusting 
all noise variables  in $S$, is given by
\begin{equation}\label{eq:iccriver}
 \var (X_{Sy}| do(\bN_{S} = \bn_S)) = \var( \sum_{j \in \bar{S}} N_j ),  
\end{equation} 
which does not coincide with the sum of the corresponding variances because the noise terms heavily correlate due to confounding. To compute the Shapley value computed for the set function \eqref{eq:iccriver}
(up to a negative sign) 
we have used variance-based feature relevance attribution
{\tt gcm.parent\_relevance} from DoWhy GCM  \citep{bloebaum2022dowhygcm} with linear regression. 

To explicitly show the difference to do-intervention based measures, we consider a modification of do-Shapley 
using variance as target metric instead of expectation based on the set function
\begin{equation}\label{eq:doshap}
\nu (S) := - \sum_{\bx_{\bar{S}}} \var (X_n| do(\bX_{S}=\bx_{S})) p(\bx_{S}),
\end{equation}
where $S \subset \{1,\dots,n-1\} $  since conditioning on $do(X_n=x_n)$ is pointless.
Whenever the mediator New Jumbles Rock is in the adjustment set, all the variance that originates from the  root nodes is screened off. 
Accordingly, New Jumbles Rock gets quite high contribution, namely around $41\%$. 
To compute variance-based do-Shapley for this example, we have used the following rule:
whenever New Jumbles Rock is in the adjustment set $S$,  we have $\var(X_{Sy}| do(\bN_{S}  = 
\bn_{S} ) = \var (N_{Sy})$, that is the influx at Samlesbury is the only source of uncertainty. 
Otherwise (that is, the adjustment set contains only root nodes), the variance of Samlesbury 
is given by the variance of the sum over all noise terms of the non-adjusted nodes and reads 
$\var(X_{Sy}| do(\bN_{S}  = 
\bn_{S} )) = \var( \sum_{j \in \bar{S}} N_j )$.   

We also tried `causal-DAG agnostic' Shapley value based feature relevance attribution 
where the interventional conditional in \eqref{eq:doshap} is replaced with the observational conditional. Then,
more than $100\%$ of the variance is attributed to the mediator $X_{JR}$ (with slightly negative contributions of the root nodes), aligning with the fact that
all other features were conditionally independent of the target, given the mediator, if there were no confounding effects.       
There, we have again used the function {\tt gcm.parent\_relevance} from DoWhy GCM  \citep{bloebaum2022dowhygcm}, but this time we regress $X_{Sy}$ on the river flows of the upstream 
nodes, rather than on the noise (i.e. the influx). 

The fact that the flow at New Jumbles Rock is mostly explained by the flow of the root nodes is only reflected by the ICC values and not the other approaches of quantifying influence mentioned here. 
\end{document}